\documentclass[10pt,twocolumn,letterpaper]{article}

\usepackage[pagenumbers]{cvpr} 
\makeatletter
\@namedef{ver@everyshi.sty}{}
\makeatother

\usepackage{graphicx}
\usepackage{amsmath}
\usepackage{mathtools}
\usepackage{amssymb}
\usepackage{amsxtra}
\usepackage{booktabs}
\usepackage[noend]{algpseudocode}

\usepackage{algorithmicx,algorithm}
\usepackage{multirow}
\usepackage{colortbl}
\usepackage{color}
\usepackage{wrapfig}
\usepackage{changes}
\usepackage{enumerate}
\usepackage{xcolor,soul,framed} 
\usepackage{float}
\usepackage{hhline}

\usepackage[pagebackref,breaklinks,colorlinks]{hyperref}

\newcommand\sbtst{\bgroup\markoverwith{\textcolor{yellow}{\rule[0.5ex]{2pt}{0.5pt}}}\ULon}

\usepackage[capitalize]{cleveref}
\crefname{section}{Sec.}{Secs.}
\Crefname{section}{Section}{Sections}
\Crefname{table}{Table}{Tables}
\crefname{table}{Tab.}{Tabs.}

\def\cvprPaperID{6173} 
\def\confName{CVPR}
\def\confYear{2023}

\begin{document}

\title{Uni3D: A Unified Baseline for Multi-dataset 3D Object Detection}

\author{Bo Zhang$^1$, Jiakang Yuan$^2$, Botian Shi$^{\dagger,1}$, Tao Chen$^2$, Yikang Li$^1$, Yu Qiao$^1$\\
$^1$ Shanghai AI Laboratory\\ $^2$School of Information Science and Technology, Fudan University\\
{\tt\small \{zhangbo, yuanjiakang, shibotian, liyikang, qiaoyu\}@pjlab.org.cn}}
\maketitle

\newcommand\blfootnote[1]{%
\begingroup
\renewcommand\thefootnote{}\footnote{#1}%
\endgroup
}
\blfootnote{{$^\dagger$}Corresponding to: Botian Shi (shibotian@pjlab.org.cn)}

\begin{abstract}
Current 3D object detection models follow a single dataset-specific training and testing paradigm, which often faces a serious detection accuracy drop when they are directly deployed in another dataset. In this paper, we study the task of training a unified 3D detector from multiple datasets. We observe that this appears to be a challenging task, which is mainly due to that these datasets present substantial data-level differences and taxonomy-level variations caused by different LiDAR types and data acquisition standards. Inspired by such observation, we present a Uni3D which leverages a simple data-level correction operation and a designed semantic-level coupling-and-recoupling module to alleviate the unavoidable data-level and taxonomy-level differences, respectively. Our method is simple and easily combined with many 3D object detection baselines such as PV-RCNN and Voxel-RCNN, enabling them to effectively learn from multiple off-the-shelf 3D datasets to obtain more discriminative and generalizable representations. Experiments are conducted on many dataset consolidation settings including Waymo-nuScenes, nuScenes-KITTI, Waymo-KITTI, and Waymo-nuScenes-KITTI consolidations. Their results demonstrate that Uni3D exceeds a series of individual detectors trained on a single dataset, with a 1.04× parameter increase over a selected baseline detector. We expect this work will inspire the research of 3D generalization since it will push the limits of perceptual performance. Our code is available at: \textit{\textcolor{teal}{\url{https://github.com/PJLab-ADG/3DTrans}}}.
\end{abstract}

\vspace{-0.2cm}
\section{Introduction}
\label{sec:intro}

\begin{figure}[t]
\centering
\includegraphics[width=6.0cm]{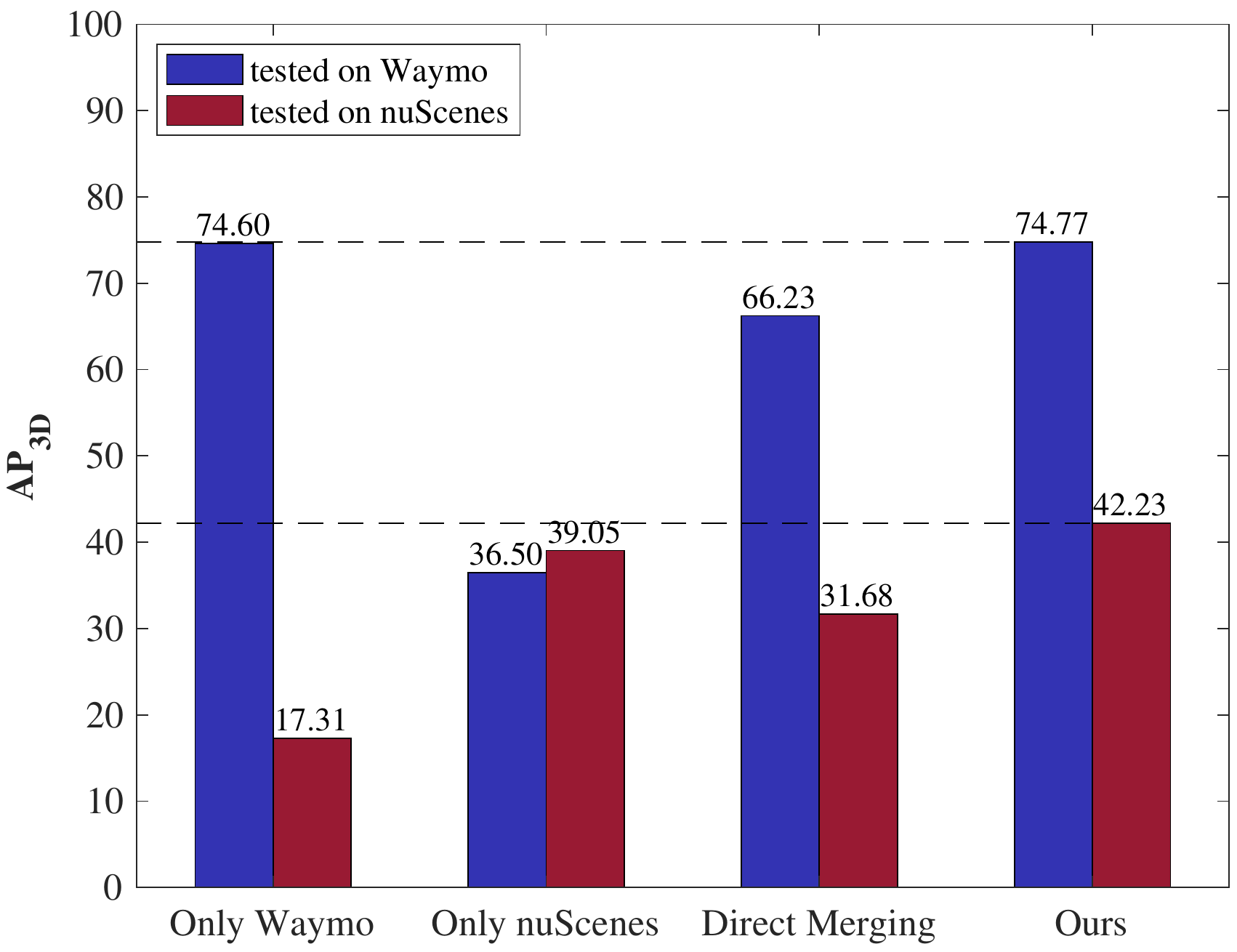}
\vspace{-0.3cm}
\caption{Challenges in training a detector from multiple datasets: 1) Only Waymo and Only nuScenes refer to the baseline detector trained on each individual dataset. 2) Direct Merging represents that we simply merge Waymo and nuScenes and train the detector on the merged dataset. 3) Ours denotes that the baseline detector is trained using the proposed method on the merged dataset. }
\label{fig1}
\vspace{-0.3cm}
\end{figure}

LiDAR-based 3D object detection~\cite{geiger2012we,wei2022lidar,shi2020pv,shi2021pv,zhou2018voxelnet,deng2021voxel,chen2017multi,lang2019pointpillars,yin2021center} aims to recognize and localize instance objects from a given frame using LiDAR sensor, which recently has achieved great progress owing to the rapid development of the large-scale annotated 3D LiDAR datasets such as Waymo~\cite{sun2020scalability}, nuScenes~\cite{caesar2020nuscenes}, and KITTI~\cite{geiger2012we}.

Unfortunately, the existing supervised 3D object detection models are designed by following the typical single-dataset training-and-testing paradigm, which inevitably suffers from severe accuracy drop issue~\cite{yang2021st3d,yang2022st3d++}, when these detection models are directly deployed to another dataset with different data distributions. For example, Fig~\ref{fig1} indicates that the baseline detector trained on Waymo~\cite{sun2020scalability} suffers from a serious detection accuracy degradation (from $74.60\%$ to $17.31\%$) when it is evaluated on another different dataset nuScenes~\cite{caesar2020nuscenes}. As a result, such a single-dataset training-and-testing paradigm cannot perform well on different datasets, further hurting the dataset-level generalization ability of the current 3D perception models.

In order to reduce the differences between different 3D datasets, some researchers~\cite{luo2021unsupervised,yang2021st3d,yang2022st3d++,wei2022lidar,xu2021spg} try to leverage Unsupervised Domain Adaptation (UDA) technique, which aims to transfer a pre-trained source-domain detector to a new domain (or dataset). Although these UDA-based 3D object detection works achieve good detection accuracy gains on the new target domain, they are still a source-to-target unidirectional model adaptation process, rather than a multi-dataset bi-directional generalization process.

Accordingly, to design a unified 3D object detection framework that can fully learn from different target datasets, we start by directly merging multiple datasets and re-training the baseline detector on the merged dataset, and found that the multi-dataset detection accuracy achieved by such a simple way is unsatisfactory, as illustrated in the results of \texttt{Direct Merging} in Fig.~\ref{fig1}. This is mainly because compared with 2D image domain, 3D point cloud data present more serious cross-dataset discrepancies caused by various and complex reasons including sensor type differences, traffic scene changes, data acquisition variations, and \textit{etc}, which is termed as \textbf{dataset-interference issue}. Further, with the constant increase of autonomous driving datasets, it becomes a very important topic for how to train a unified detector from such diversified 3D datasets. 

In this paper, we propose a Unified 3D object detection framework (Uni3D) to address the dataset-interference issue. Orthogonal to the existing 3D object detection research works~\cite{shi2020pv,deng2021voxel,lang2019pointpillars,shi2019pointrcnn,yan2018second} focusing on developing an effective framework verified within a single dataset, Uni3D aims to propose a simple-and-versatile way to enable the existing 3D object detection models to have the ability of learning from many diversified 3D datasets. To achieve this goal, we design a simple data-level correction operation that can use dataset-specific channel-wise mean and variance to normalize features from each backbone layer. Besides, a semantic-level coupling-and-recoupling module is designed to strengthen the feature reusability across different 3D datasets, by calculating a spatial-wise attention map and a dataset-level attention mask to constrain the learned high-level features to be dataset-agnostic. 

Extensive experiments are conducted on three public 3D autonomous driving datasets including Waymo~\cite{sun2020scalability}, nuScenes~\cite{caesar2020nuscenes}, and KITTI~\cite{geiger2012we}, to investigate the reasons for the 3D dataset-interference issue. Besides, this paper provides many preliminary studies that explore the possibility of training a 3D object detection model under the merged datasets. The experimental results show that Uni3D has a strong dataset-level generalization ability, improving the zero-shot learning ability for unseen scenes and even surpassing the baseline methods trained on a single dataset.

\vspace{-0.15cm}
\section{Related Works}
\vspace{-0.1cm}
\subsection{LiDAR-based General 3D Object Detection}
\vspace{-0.15cm}
Recent LiDAR-based 3D object detection works~\cite{chen2017multi, deng2021voxel, lang2019pointpillars, qi2018frustum, shi2019pointrcnn, shi2020points, shi2020pv, yan2018second,yang2018pixor, yang2019std, zhou2018voxelnet} can be roughly categorized into voxel-based methods, point-based methods, and point-voxel fusion methods. Voxel-based methods~\cite{sindagi2019mvx,yan2018second,zhou2018voxelnet} convert irregular LiDAR points to ordered voxels before backbone feature extraction. SECOND~\cite{yan2018second} is a prior work that utilizes sparse convolution as 3D backbone and greatly improves the detection efficiency. Voxel-RCNN~\cite{deng2021voxel} analyses the advantages of voxel features and explores a good trade-off between detection accuracy and inference speed. Unlike voxel-based methods, point-based methods~\cite{zhang2022not,shi2019pointrcnn} directly generate feature maps from raw point clouds. Inspired by PointNet~\cite{qi2017pointnet} and PointNet++~\cite{qi2017pointnet++}, Point-RCNN~\cite{shi2019pointrcnn} is a pioneer to investigate how to generate bounding boxes from point cloud data. To reduce the high memory and computational cost of point-based methods, IA-SSD~\cite{zhang2022not} proposes a single-stage method by employing learning-based instance-aware down-sampling strategies. Besides, some works try to combine the benefits of point- and voxel-based representations. Among them, PV-RCNN~\cite{shi2020pv} designs a point-voxel feature set abstraction to fully combine point features and voxel features. However, the above detectors are trained and evaluated within a single 3D dataset, and they will suffer from severe detection accuracy drop issues across different datasets. Further, learning generalizable representations between different datasets is more challenging in 3D scenarios due to the more serious dataset-level gaps. 

\vspace{-0.10cm}
\subsection{Joint Training on Multiple Datasets} 
\vspace{-0.10cm}
For traditional 2D perception tasks such as object detection~\cite{zhu2020deformable,ren2015faster} and semantic segmentation~\cite{zheng2021rethinking}, training a unified model from different datasets results in a low recognition accuracy, since different datasets often present inconsistent class definitions and annotation granularity. Motivated by this, some researchers start to study how to achieve a multi-dataset perception task~\cite{wang2019towards,zhou2022simple,lambert2020mseg,zhao2020object,dai2021dynamic,gong2021mdalu}. Early works~\cite{lambert2020mseg,zhao2020object} focus on merging the taxonomy information and train the model on a unified label space. Mseg~\cite{lambert2020mseg} aligns pixel-level annotations of seven datasets and significantly boosts the generalization ability of the model. Zhao \etal~\cite{zhao2020object} propose to train a dataset-specific detector to generate pseudo labels, which provide additional annotation information from another dataset, and the final network is trained on a specific dataset using both pseudo labels and ground truths. To alleviate the annotation cost of unifying the label space, recent works~\cite{wang2019towards,zhou2022simple} attempt to use dataset-specific supervision. Wang~\etal~\cite{wang2019towards} employ a designed domain adaptation layer and attention mechanism to alleviate the dataset-level differences. Zhou~\etal~\cite{zhou2022simple} introduce a novel automatic way to merge the taxonomy space, showing that the unified detector trained on multiple datasets can outperform each detector trained on the specific dataset. Although jointly training a unified detector has been recently studied in 2D perception tasks, its further exploration on 3D perception tasks, such as 3D object detection, is still insufficient.

\vspace{-0.15cm}
\section{The Proposed Method}
\label{sec:method}
\vspace{-0.10cm}
The overall framework is shown in Fig.~\ref{fig:framework}. We first describe our problem setting and the multi-dataset evaluation method. Next, we analyze the limitations of the current baseline detector in multi-dataset detection, and then introduce a simple solution, namely, Uni3D.

\subsection{Preliminary}
\vspace{-0.15cm}
\noindent\textbf{Problem Setting.} Suppose that a domain is defined by a joint probability distribution $P_{XY}$ on $\mathcal{X} \times \mathcal{Y}$, where $\mathcal{X}$ and $\mathcal{Y}$ are the input point cloud and label space, respectively. In the scope of \textbf{Multi-Domain Fusion (MDF)}, $N$ denotes the number of domains  $\mathcal{S}=\left\{S_n=\left\{\left(\mathbf{x}^{(n)}, y^{(n}\right)\right\}\right\}_{n=1}^N$ available for model training, where each individual domain $S_n$ is associated with a specific data distribution $P_{XY}^{n}$. The purpose of MDF is to train a unified model from multiple labeled domains $\mathcal{S}$ to obtain more generalizable representations $F: \mathcal{X} \rightarrow  \mathcal{Y}$, which would have minimum prediction error on the multiple different domains $\mathcal{S}$. 

\noindent\textbf{3D Multi-dataset Training and Evaluation.} Assume that in real application, we can simultaneously access multiple labeled 3D point cloud-based domains or datasets (\textit{e.g.,} Waymo~\cite{sun2020scalability} and nuScenes~\cite{caesar2020nuscenes}), but these labeled datasets often have different label space $\mathcal{Y}$, such as Barrier category which only presents in nuScenes~\cite{caesar2020nuscenes}. Our study mainly focuses on MDF under autonomous driving scenario, where the model training and evaluation are conducted on the categories of interest related to autonomous driving scenario: vehicle, pedestrian, and cyclist. Note that \textbf{such a setting of selecting common categories, such as vehicle, pedestrian, and cyclist categories, to conduct the preliminary research is very common in many cross-dataset 3D detection works}, such as ST3D~\cite{yang2021st3d}, ST3D++~\cite{yang2022st3d++}.

\subsection{When Single-dataset 3D Detectors Meet Multiple Datasets}
\vspace{-0.10cm}

\noindent\textbf{Single-dataset 3D Object Detection.}
Currently, state-of-the-art 3D object detection models~\cite{shi2020points,yan2018second,shi2020pv,shi2021pv,deng2021voxel,lang2019pointpillars} are trained and evaluated within a single public benchmark~\cite{sun2020scalability,caesar2020nuscenes}, which can be regarded as so-called \textbf{single-dataset training paradigm}. Here, to better illustrate our MDF task, we first abstract the optimization objective of current 3D object detection models as follows:
\vspace{-0.10cm}
\begin{equation}
    L_{det}=L_{rpn}+L_{roi}+L_{key},
    \label{eq:pvrcnn_loss}
\end{equation}

\vspace{-0.20cm}
\noindent where {\small $L_{rpn}$} is used to generate accurate localization prediction of preset proposals, and {\small $L_{roi}$} helps to refine the proposals to obtain the final 3D bounding box results. Besides, some 3D baseline detectors, such as PV-RCNN~\cite{shi2020pv} and PV-RCNN++~\cite{shi2021pv}, use a keypoint prediction loss {\small $L_{key}$} that can identify important foreground points and achieve a keypoint-to-grid RoI feature extraction process. 

\noindent\textbf{Limitation for MDF-based 3D Object Detection.} 
A natural method to train on multiple datasets is to simply combine all source datasets into a merged but larger one. Unfortunately, our initial attempt of such a natural dataset combination way shows a significant performance drop of the detector that is trained on the merged dataset, compared with the performance of training on each specific sub-dataset. 

As demonstrated in Table~\ref{tab3} of Sec.~\ref{sec4.3}, we observe that, by comparing \texttt{Single-dataset} and \texttt{Direct Merging} baselines, for 3D scene-level datasets, directly perform a dataset-level consolidation cannot help to boost the detector's cross-dataset detection accuracy. On the contrary, the detector may suffer from the feature learning interference due to the significant differences between different datasets. For example, Voxel-RCNN~\cite{deng2021voxel} can obtain a relatively-high detection accuracy (75.08$\%$ AP on Waymo validation set) when it is trained only on Waymo~\cite{sun2020scalability} dataset. But it faces a severe performance drop when Voxel-RCNN is jointly trained on the combined dataset of Waymo and nuScenes (only 66.67$\%$ AP on Waymo validation set).   

Here, through extensive experiments, we give two main reasons for the above performance degradation issue.

\begin{figure}[t]
\vspace{-0.30cm}
\centering
\includegraphics[width=7.5cm,height=4.1cm]{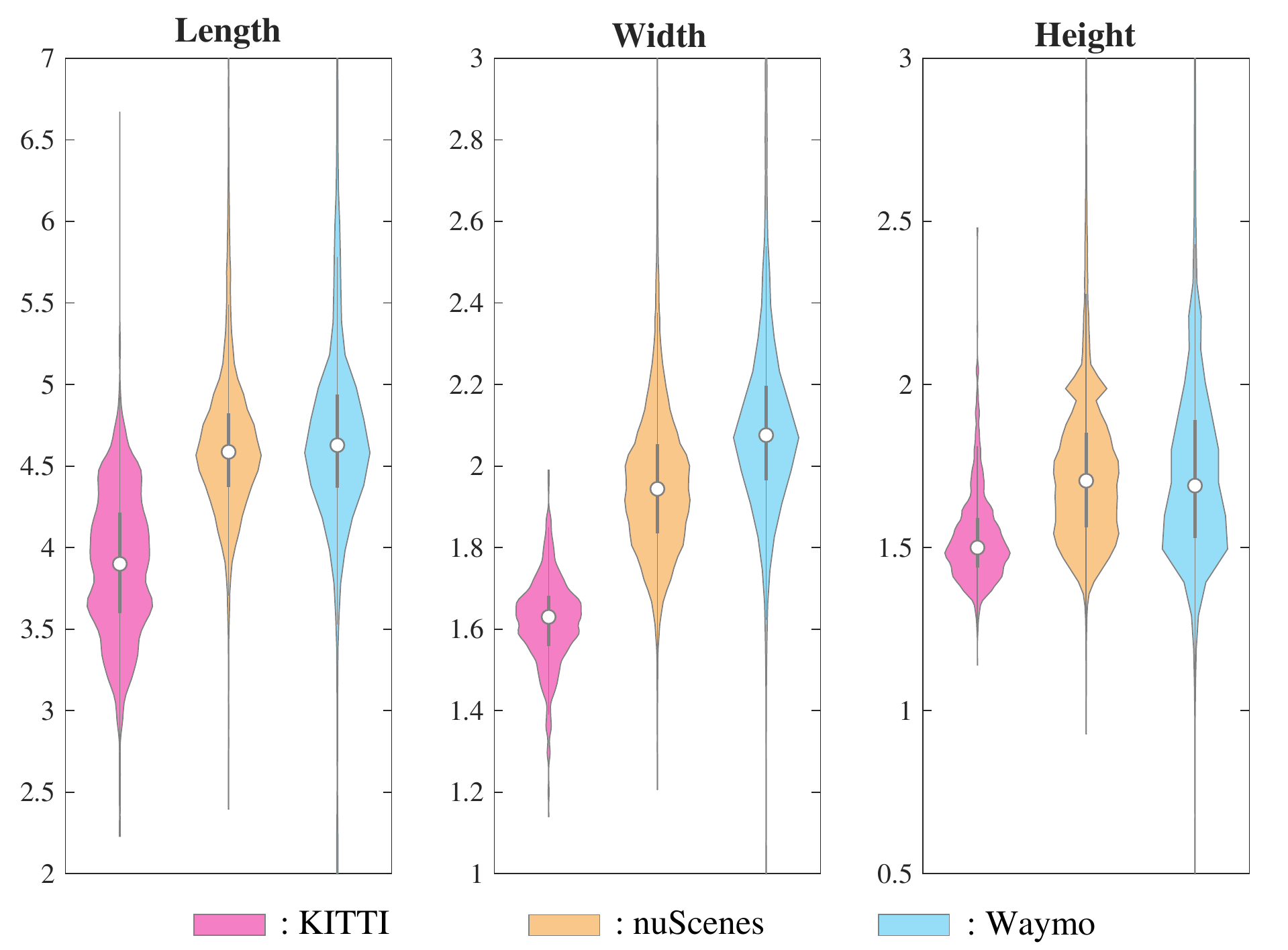}
\vspace{-0.3cm}
\caption{The statistical distribution differences of object size (Length, Width, and Height) across different datasets. To better illustrate the differences, we pick up the values within the range of [2.0, 7.0], [1.0, 3.0], and [0.5, 3.0] for Length, Width, and Height.}
\label{fig2}
\vspace{-0.10cm}
\end{figure}

\begin{table}[t]
\small
\centering
\setlength{\tabcolsep}{0.5mm}{
\resizebox{\linewidth}{!}
{
\begin{tabular}{ccc|c|c}
\hline
Datasets & Beam & VFOV & Point Range & Collection Location\\
\hline
\multirow{3}{*}{Waymo~\cite{sun2020scalability}} & \multirow{3}{*}{64} &  \multirow{3}{*}{[$-18.0^\circ$, $2.0^\circ$]} & L=[-75.2, 75.2]m & \multirow{3}{*}{USA} \\
&  &  & W=[-75.2, 75.2]m  &  \\
&  &  & H=[-2.0, 4.0]m  &  \\
\hline
\multirow{3}{*}{KITTI~\cite{geiger2012we}} & \multirow{3}{*}{64} & \multirow{3}{*}{[$-23.6^\circ$, $3.2^\circ$]} & L=[0.0, 70.4]m & \multirow{3}{*}{Germany} \\
&  &  & W=[-40.0, 40.0]m  &  \\
&  &  & H=[-3.0, 1.0]m  &  \\
\hline
\multirow{3}{*}{nuScenes~\cite{caesar2020nuscenes}} & \multirow{3}{*}{32} & \multirow{3}{*}{[$-30.0^\circ$, $10.0^\circ$]} & L=[-51.2, 51.2]m  & \multirow{3}{*}{USA/Singapore} \\
&  &  & W=[-51.2, 51.2]m  &  \\
&  &  & H=[-5.0, 3.0]m  &  \\
\bottomrule
\end{tabular}
}
}
\vspace{-0.3cm}
\caption{Overview of 3D autonomous driving dataset differences. VFOV denotes vertical field of view, and L, W, and H represent the length, width, and height of LiDAR range, respectively.}
\label{tab1}
\vspace{-0.35cm}
\end{table}

\begin{figure*}[t]
\vspace{-0.4cm}
\centering
\includegraphics[width=16.5cm,height=5.6cm]{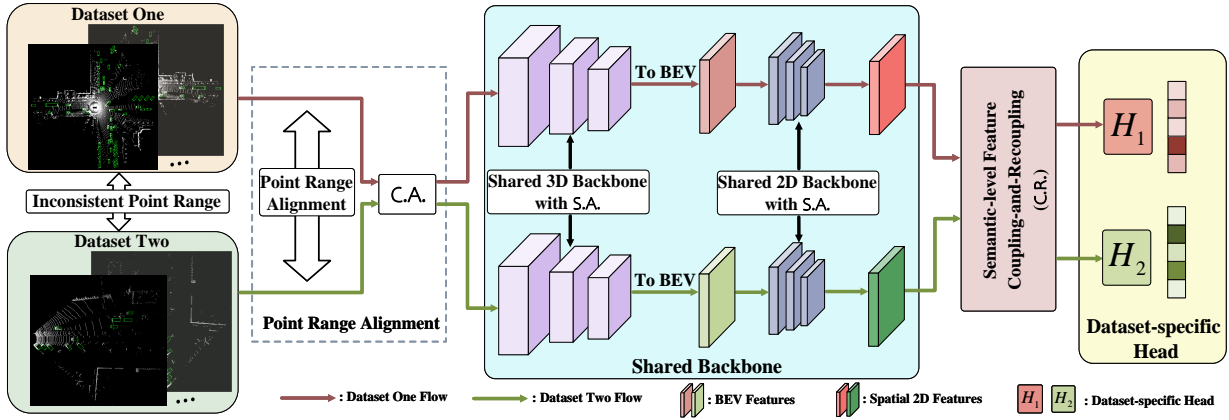}
\vspace{-0.3cm}
\caption{The overview of Uni3D including: 1) point range alignment, 2) parameter-shared 3D and 2D backbones with data-level correction operation, 3) semantic-level feature coupling-and-recoupling module, and 4) dataset-specific detection heads. \texttt{C.A.} denotes Coordinate-origin Alignment to reduce the adverse effects caused by point range alignment, and \texttt{S.A.} is the designed Statistics-level Alignment.}
\vspace{-0.2cm}
\label{fig:framework}
\end{figure*}

\noindent \underline{\textbf{1) Data-level differences}}: Compared with 2D natural images that are composed of pixels with a consistent value range of [0, 255], 3D point clouds often are collected using different sensor types with different point cloud ranges, which leads to distributional discrepancy among datasets. And the main differences of the three widely-used datasets are shown in Table~\ref{tab1}. Actually, we found from Table~\ref{tab2} that sensor-derived point range difference is a major factor interfering the common feature learning from multiple datasets, which is due to that the receptive field size for the same objects are very different when data with inconsistent point cloud ranges are fed into the 3D detector. As a result, the point-cloud-range alignment is a necessary pre-processing step for achieving multi-dataset 3D object detection.

\begin{table}[]
\vspace{-0.05cm}
\small
\centering
\setlength{\tabcolsep}{1.8mm}{
\resizebox{\linewidth}{!}
{
\begin{tabular}{ccccc}
\hline
\multirow{2}{*}{Methods} & \multirow{2}{*}{Waymo Range} & \multirow{2}{*}{KITTI Range} & tested on Waymo & tested on KITTI \\
 &  &  & $\text{AP}_{\text{3d}}$ / $\text{APH}_{\text{3D}}$ &  $\text{AP}_{\text{BEV}}$ / $\text{AP}_{\text{3D}}$  \\ 
\hline
\multirow{3}{*}{Not Align.} & L=[-75.2, 75.2]m & L=[0.0, 70.4]m & \multirow{3}{*}{26.93 / 26.56} & \multirow{3}{*}{89.56 / \textbf{83.14}} \\
& W=[-75.2, 75.2]m & W=[-40.0, 40.0]m  & \\
& H=[-2.0, 4.0]m & H=[-3.0, 1.0]m & \\
\toprule[0.5pt]
\multirow{3}{*}{Align. (w/ ours)} & L=[-75.2, 75.2]m & L=[-75.2, 75.2]m & \multirow{3}{*}{\textbf{74.83} / \textbf{74.33}} & \multirow{3}{*}{\textbf{90.03} / 82.39} \\
& W=[-75.2, 75.2]m & W=[-75.2, 75.2]m  & \\
& H=[-2.0, 4.0]m & H=[-2.0, 4.0]m & \\
\toprule[0.5pt]
\toprule[0.5pt]
\multirow{2}{*}{Methods} & \multirow{2}{*}{nuScenes Range} & \multirow{2}{*}{KITTI Range} & tested on nuScenes & tested on KITTI \\
 &  &  & $\text{AP}_{\text{BEV}}$ / $\text{AP}_{\text{3D}}$  &  $\text{AP}_{\text{BEV}}$ / $\text{AP}_{\text{3D}}$  \\ 
\hline
\multirow{3}{*}{Not Align.} & L=[-51.2, 51.2]m & L=[0.0, 70.4]m & \multirow{3}{*}{21.32 / 15.35} & \multirow{3}{*}{89.35 / 81.66} \\
& W=[-51.2, 51.2]m & W=[-40.0, 40.0]m  & \\
& H=[-5.0, 3.0]m & H=[-3.0, 1.0]m & \\
\toprule[0.5pt]
\multirow{3}{*}{Align. (w/ ours)} & L=[-75.2, 75.2]m & L=[-75.2, 75.2]m & \multirow{3}{*}{\textbf{59.25} / \textbf{41.51}} & \multirow{3}{*}{\textbf{90.09} / \textbf{83.10}} \\
& W=[-75.2, 75.2]m & W=[-75.2, 75.2]m  & \\
& H=[-2.0, 4.0]m & H=[-2.0, 4.0]m & \\
\hline
\end{tabular}}
}
\vspace{-0.3cm}
\caption{Inconsistent LiDAR ranges will cause the multi-dataset detection accuracy drop. The baseline employs Voxel-RCNN~\cite{deng2021voxel}, and please refer to Appendix for all-category results.}
\label{tab2}
\vspace{-0.5cm}
\end{table}

Besides, as illustrated in Fig.~\ref{fig2}, the point clouds from different datasets present a more diverse data distribution, due to that 3D datasets were collected in different cities and countries whose instance size is very different.

\noindent \underline{\textbf{2) Taxonomy-level differences}}: 
Given a fact that different autonomous driving manufacturers employ inconsistent class definitions and annotation granularity. For example, for Waymo~\cite{sun2020scalability}, all vehicles driving on the road, including car and truck, are annotated as one unified category, namely, `Vehicle'. While for nuScenes~\cite{caesar2020nuscenes}, different vehicles are annotated using different taxonomies with different granularity, such as `Car', `Truck, and `Van'. As a result, the MDF task needs to consider how to train a 3D detector under an inconsistent taxonomy label space and effectively reuse domain-agnostic knowledge that can be shared across different datasets.

\subsection{Uni3D: A Unified 3D Multi-dataset Object Detection Baseline}
\label{subsec:method}
\vspace{-0.15cm}

To address the data- and taxonomy-level difference issue described in the last section, we aim to develop simple modules that enable the existing 3D detectors~\cite{shi2020pv, deng2021voxel} to learn generalizable representations from different datasets.

\noindent\textbf{Data-level Correction Operation.} Firstly, to accomplish a data-level correction operation, we introduce a Statistic-level Alignment (\texttt{S.A.}) that can alleviate the statistic-level differences of features extracted by common 2D or 3D backbones. This approach 
can be combined with any 3D detectors, including PV-RCNN~\cite{shi2020pv} and Voxel-RCNN~\cite{deng2021voxel}.

Specifically, suppose that $\mu^j$ and $\sigma^j$ denote the mean and variance for each channel in the $j$-th network layer. Generally, the basic mean and variance statistics are used to normalize the feature in each layer, such that the input data for each layer comply with zero-mean and univariance, \textbf{e.g.,} BN~\cite{ioffe2015batch}. However, such a statistics-shared normalization way may hurt the model transferability in MDF training, since data within one batch-size may come from different datasets having large mean and variance differences. To this end, under the MDF setting, we first obtain dataset-specific channel-wise mean $\mu^j_t$ and $\sigma^j_t$ variance from the $t$-th dataset. Then, samples from each dataset are regularized by the current dataset-specific mean/variance as follows:
\vspace{-0.10cm}
\begin{equation}
    \label{eq:center}
    \hat{x}_t^j \ = \ \frac{x_t^j - \mu^j_t}{\sqrt{\sigma_t^j + \xi}},
\end{equation}
\vspace{-0.10cm}

\noindent where $x_t^j$ denotes the input feature of each network layer in the $t$-th dataset, and $\xi$ is added to ensure the numerical stability. Further, similar to the BN~\cite{ioffe2015batch}, a transformation step is employed to restore the representation ability as follows:
\begin{equation}
    \label{eq:center}
    \hat{y}_t^j \ = \ \gamma^j \hat{x}_t^j + \beta^j,
\end{equation}

\vspace{-0.10cm}
\noindent where we employ the dataset-shared gamma $\gamma$ and beta $\beta$, since after the features from different datasets are normalized to zero-mean and univariance, the inter-dataset differences in the first/second-order are aligned, and further, it is reasonable to use the same gamma $\gamma$ and beta $\beta$ for different datasets.

\noindent\textbf{Semantic-level Feature Coupling-and-Recoupling Module.} In this part, we introduce a simple semantic-level Coupling-and-Recoupling module (\texttt{C.R.}) that also can be easily inserted into many single-dataset 3D detectors, to exploit the reusable features across datasets from two aspects: 1) Feature Coupling and 2) Feature Recoupling.

\noindent\textit{1) Feature Coupling}: Suppose that {\small $f^{bev}\in\mathbb{R}^{C\times H \times W}$} denotes the Bird-Eye-View (BEV) features extracted by 2D backbone network, where {\small $C$} denotes the channel number, {\small $H$} and {\small$W$} are the height and width of the features, respectively. BEV features {\small $f^{bev}\in\mathbb{R}^{C\times H \times W}$} from different datasets are coupled together along the channel dimension to learn dataset-agnostic representations using a foreground-aware and dataset-level attention mask as follows:
\vspace{-0.35cm}

\begin{equation}
\begin{split}
\begin{aligned}
&f_{cat}^{bev} \ = \ [f_i^{bev}, \ ..., \ \ f_j^{bev}], \\
&\hat{f}^{bev}_{shared}=  [M_{shared} \ \odot \ \phi_{d}(Conv(f_{cat}^{bev}))] \ f_{cat}^{bev}, 
\end{aligned}
\end{split}
\end{equation}

\vspace{-0.15cm}
\noindent where {\small $[..., ...]$} is the concatenation operation along the channel dimension, {\small $f_i^{bev}$} and {\small $f_j^{bev}$} represent the BEV features from the $i$-th and $j$-th dataset, respectively. And {\small$M_{shared} = \phi_{p}(f_{cat}^{bev})$}, where {\small $\phi_{p}$} denotes the foreground-aware spatial attention operation which is achieved by calculating the channel-wise maximum value of BEV features. Besides, the {\small $\phi_{d}$} represents the dataset-level attention mask achieved by a Multiple Layer Perceptron (MLP) {\small $Conv(f_{cat}^{bev})$} followed by a $N$-cls softmax operation, where $N$ denotes the number of datasets to be merged. Such a dataset-level attention mask means that the {\small $\phi_{d}$} can predict a re-scaling score to recombine BEV features from different datasets so that the combined BEV features are dataset-agnostic. 

\noindent\textit{2) Feature Recoupling}: Since the shared features {\small $\hat{f}^{bev}_{shared}$} mainly focus on the common knowledge of multiple datasets, we are expected to fuse such shared features {\small $\hat{f}^{bev}_{shared}$} with previous dataset-related BEV features {\small $f^{bev}_{i}$} or {\small $f^{bev}_{j}$} using a channel-wise re-scaling operation:
\vspace{-0.35cm}

\begin{equation}
\begin{split}
&\hat{f}_{i}^{bev} \ = \ SE_i(\hat{f}_{shared}^{bev}) \ + \ f_{i}^{bev}, \\
&\hat{f}_{j}^{bev} \ = \ SE_j(\hat{f}_{shared}^{bev}) \ + \ f_{j}^{bev},  
\end{split}
\end{equation}

\begin{figure}[t]
\vspace{-0.35cm}
\centering
\includegraphics[width=7.2cm]{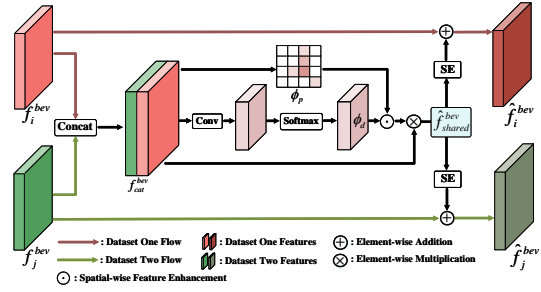}
\vspace{-0.30cm}
\caption{Semantic-level feature coupling-and-recoupling during the multi-dataset training stage.}
\label{fig:C.R.}
\vspace{-0.40cm}
\end{figure}

\vspace{-0.15cm}
\noindent where {\small $SE$} is the Squeeze-and-Excitation Network~\cite{hu2018squeeze}, and its network architecture is described in Appendix. The overall network structure of the designed coupling-and-recoupling module is illustrated in Fig.~\ref{fig:C.R.}. However, exploring such feature relations across datasets will cause the inconsistency between the model multi-dataset training and single-dataset testing, mainly due to that the shared features {\small $\hat{f}^{bev}_{shared}$} are dependent on multiple inputs. To tackle this issue, we simply use the \underline{\textbf{BEV feature copy}} method, meaning that during the single-dataset inference stage, BEV features from the single dataset will be simultaneously copied to $f_i^{bev}$ and $f_j^{bev}$, to obtain the shared features {\small $\hat{f}^{bev}_{shared}$}. As a result, BEV feature copy method enables us to perform the inference on a single dataset, and Uni3D is not depending on some given frame from another domain during the inference. Ablation studies on different training-and-testing methods (including \textbf{BEV feature copy} and \textbf{BEV feature mask}) are described in Appendix.

\noindent\textbf{Dataset-specific Detection Heads.} To further address the taxonomy-level differences between datasets, we assume that the prior knowledge of 3D data sources is known, and propose to use different detection heads $H_k$ followed by dataset-specific detection loss $L_{det}^k$ to perform the instance-level prediction on different datasets. The MDF training loss function {\small $L_{det}^{overall}$} can be written as follows:

\vspace{-0.3cm}
\begin{equation}
\begin{split}
L_{det}^{overall}= \ \sum_{\boldsymbol{k}} \ L_{det}^k (H_k(\hat{f}_k^{bev})),
\end{split}
\end{equation}

\vspace{-0.2cm}
\noindent where {\small $L_{det}^k$} is the dataset-specific loss from the $k$-th dataset.

During the inference phase, we use the data-level correction operation to address the sensor-induced or data-distribution-induced differences, and the parameter-shared 3D and 2D backbones to extract the point and voxel features. Meanwhile, the detection head assigned to the corresponding dataset is used to produce final prediction results.

\begin{table*}[t]
\vspace{-0.4cm}
    \centering
    \begin{small}
    \resizebox{0.94\linewidth}{!}
    {
        \begin{tabular}{c|c|c|c|c|c|c|c}
            \bottomrule[1pt]
            \multirow{2}{*}{Trained on} & \multirow{2}{*}{Baseline Detectors}  & \multicolumn{3}{c|}{Tested on Waymo} & \multicolumn{3}{c}{Tested on nuScenes} \\
            \hhline{~~|-|-|-|-|-|-}
            &  & Vehicle & Pedestrian & Cyclist & Car & Pedestrian & Cyclist \\
            \hline
            \multirow{2}{*}{only Waymo} & Voxel-RCNN~\cite{deng2021voxel} (w/o \texttt{P.T.})  & 75.08 / 74.60 & 75.17 / \textbf{68.76} &  65.28 / 64.33 &  34.10 / 17.31 & 2.99 / 1.69 & 0.05 / 0.01 \\
            & Voxel-RCNN~\cite{deng2021voxel} (w/ \texttt{P.T.} on nuScenes) & \textbf{75.46} / \textbf{74.99} & 74.58 / 68.06 & \textbf{65.92} / \textbf{64.98} & 34.34 / 21.95 & 2.84 / 1.57 & 0.09 / 0.02  \\
            \toprule[1pt]
            \multirow{2}{*}{only nuScenes} & Voxel-RCNN~\cite{deng2021voxel} (w/o \texttt{P.T.})   & 36.77 / 36.50 & 4.64 / 3.18 &  2.49 / 2.45 &  53.63 / 39.05 & 22.47 / 17.85 & 10.86 / 9.70 \\
            & Voxel-RCNN~\cite{deng2021voxel} (w/ \texttt{P.T.} on Waymo) & 6.11 / 5.90 & 0.77 / 0.56 & 0.01 / 0.01 & 55.23 / 39.14 & 23.65 / 16.47 & 8.51 / 5.80  \\
            \toprule[1pt]
            \multirow{4}{*}{Waymo+nuScenes} & Voxel-RCNN~\cite{deng2021voxel} (w/ \texttt{D.M.})  & 66.67 / 66.23 & 60.36 / 54.08 &  52.03 / 51.25 &   51.40 / 31.68 & 15.04 / 9.99 & 5.40 / 3.87 \\
            & Voxel-RCNN~\cite{deng2021voxel} (w/ \texttt{C.A.})  & 69.40 / 68.86 & 63.43 / 56.49 & 52.83 / 51.93 & 51.39 / 29.04 & 16.24 / 10.96 & 4.55 / 3.13  \\
            & Voxel-RCNN~\cite{deng2021voxel} (w/ \texttt{C.A.+S.A.})  & 75.16 / 74.67 & 74.83 / 68.07 & 64.68 / 63.73 & 58.41 / 40.84 & 26.52 / 20.98 & 9.19 / 7.65  \\
            & Voxel-RCNN~\cite{deng2021voxel} (w/ \texttt{C.A.+C.R.})  & 74.56 / 74.05 &  74.29 / 67.04 & 63.14 / 62.21 & 59.10 / \textbf{42.25} & 29.86 / 23.76 & 14.46 / \textbf{12.73}  \\
            & Voxel-RCNN~\cite{deng2021voxel} (w/ \texttt{C.A.+S.A.+C.R.})  & 75.26 / 74.77 & \textbf{75.46} / 68.75 & 65.02 / 63.12 & \textbf{60.18} / 42.23 & \textbf{30.08} / \textbf{24.37} & \textbf{14.60} / 12.32  \\
            \toprule[1pt]
            \toprule[1pt]
            \multirow{2}{*}{only Waymo} & PV-RCNN~\cite{shi2020pv} (w/o \texttt{P.T.})   & 74.97 / 74.46 & 73.41 / 66.57 &  \textbf{64.58} / \textbf{63.49} &  32.99 / 17.55 & 3.34 / 1.94 & 0.02 / 0.01 \\
            & PV-RCNN~\cite{shi2020pv} (w/ \texttt{P.T.} on nuScenes) & 74.77 / 74.26 & 73.32 / 66.31 & 64.06 / 63.05 & 33.86 / 17.47  & 2.88 / 1.53  & 0.04 / 0.01  \\
            \toprule[1pt]
            \multirow{2}{*}{only nuScenes} & PV-RCNN~\cite{shi2020pv} (w/o \texttt{P.T.})  & 41.01 / 40.58 & 4.57 / 2.96 &  0.98 / 0.95 &   57.78 / 41.10 & 24.52 / 18.56 & 10.24 / 8.25 \\
            & PV-RCNN~\cite{shi2020pv} (w/ \texttt{P.T.} on Waymo) & 44.59 / 44.24  & 7.67 / 6.33 & 8.77 / 8.58 & 57.92 / 41.53 & 24.32 / 17.31 &  11.52 / 9.19  \\
            \toprule[1pt]
            \multirow{4}{*}{Waymo+nuScenes} & PV-RCNN~\cite{shi2020pv} (w/ \texttt{D.M.})  & 66.22 / 65.75 & 55.41 / 49.29 &  56.50 / 55.48 &   48.67 / 30.43  & 12.66 / 8.12 & 1.67 / 1.04 \\
            & PV-RCNN~\cite{shi2020pv} (w/ \texttt{C.A.})  & 66.90 / 65.61 & 56.41 / 51.06 & 56.00 / 55.00 & 48.93 / 31.21 & 14.47 / 10.31 & 1.70 / 1.07  \\
            & PV-RCNN~\cite{shi2020pv} (w/ \texttt{C.A.+S.A})  & 74.24 / 73.71 & 67.38 / 60.79  &  60.20 / 59.16  & 59.49 / 42.05 & \textbf{27.44} / 20.94 & 12.69 / 10.34  \\
            & PV-RCNN~\cite{shi2020pv} (w/ \texttt{C.A.+C.R.})  & 74.88 / 74.36 &   73.39 / 66.02 & 62.84 / 61.79 & 59.01 / 41.16 & 26.59 / 20.49 &  9.86 / 7.60  \\
            & PV-RCNN~\cite{shi2020pv} (w/ \texttt{C.A.+S.A.+C.R.})  & \textbf{75.54} / \textbf{74.90} & \textbf{74.12} / \textbf{66.90} & 63.28 / 62.12 & \textbf{60.77} / \textbf{42.66} & \textbf{27.44} / \textbf{21.85} & \textbf{13.50} / \textbf{11.87}  \\
            \toprule[1pt]
        \end{tabular}
    }
    \end{small}
    \vspace{-0.3cm}
    \caption{Results of joint training on Waymo and nuScenes datasets. Following the existing 3D object detection works~\cite{yang2021st3d,shi2020pv,yang2022st3d++}, we report the car (Vehicle on Waymo), pedestrian, and cyclist results under IoU threshold of 0.7, 0.5, and 0.5, respectively, and utilize $\text{AP}$ and $\text{APH}$ of LEVEL 1 metric on Waymo, and $\text{AP}_{\text{BEV}}$ and $\text{AP}_{\text{3D}}$ over 40 recall positions on nuScenes. The best detection results are marked using \textbf{bold}. Due to the page limitation, the average accuracy of multiple datasets is reported in Appendix.}
    \label{tab3}
\end{table*}

\begin{table*}[htbp]
    \centering
    \begin{small}
    \resizebox{0.94\linewidth}{!}
    {
        \begin{tabular}{c|c|c|c|c|c|c|c}
            \bottomrule[1pt]
            \multirow{2}{*}{Trained on} & \multirow{2}{*}{Baseline Detectors}  & \multicolumn{3}{c|}{Tested on KITTI} & \multicolumn{3}{c}{Tested on nuScenes} \\
            \hhline{~~|-|-|-|-|-|-}
            &  & Car & Pedestrian & Cyclist & Car & Pedestrian & Cyclist \\
            \hline
            \multirow{2}{*}{only KITTI} & Voxel-RCNN~\cite{deng2021voxel} (w/o \texttt{P.T.})  & 89.34 / 80.91 & 59.67 / 56.88 &  61.10 / 60.49 &  11.37 / 4.64 & 0.15 / 0.11 &  0.01 / 0.00 \\
            & Voxel-RCNN~\cite{deng2021voxel} (w/ \texttt{P.T.} on nuScenes) & 89.90 / 81.25 & 59.49 / 56.17 & 54.55 / 54.15 & 12.89 / 5.52  & 0.24 / 0.18 & 0.05 / 0.03 \\
            \toprule[1pt]
            \multirow{2}{*}{only nuScenes} & Voxel-RCNN~\cite{deng2021voxel} (w/o \texttt{P.T.})  & 69.41 / 33.48 & 28.06 / 19.20 &  0.44 / 0.43 &  53.63 / 39.05 & 22.47 / 17.85 & 10.86 / 9.70 \\
            & Voxel-RCNN~\cite{deng2021voxel} (w/ \texttt{P.T.} on KITTI) & 71.61 / 40.64 & 39.67 / 29.99 & 7.29 / 6.88 & 53.57 / 39.65 & 24.93 / 21.17 & 11.42 / 9.95  \\
            \toprule[1pt]
            \multirow{4}{*}{KITTI+nuScenes} & Voxel-RCNN~\cite{deng2021voxel} (w/ \texttt{D.M.})  &  89.24 / 73.72 & 61.03 / 54.55 &  62.71 / 59.92 &   41.88 / 20.48 &  12.58 / 8.32 & 1.77 / 0.97 \\
            & Voxel-RCNN~\cite{deng2021voxel} (w/ \texttt{C.A.})  & 89.35 / 76.77 & 59.01 / 53.67 & 43.45 / 42.41 & 49.95 / 28.43 & 16.63 / 11.93 & 3.84 / 3.12  \\
            & Voxel-RCNN~\cite{deng2021voxel} (w/ \texttt{S.A.})  & 89.21 / 82.68 & 62.32 / 57.99 &  63.10 / 61.67 & 57.87 / 40.23 &  27.21 / 21.44 &  13.65 / 12.24  \\
            & Voxel-RCNN~\cite{deng2021voxel} (w/ \texttt{C.R.})  & 89.13 / 82.50 &  61.45 / 56.65 & 61.72 / 58.66 & 58.13 / 40.26 &  27.27 / 21.50 &  13.81 / 12.18  \\
            & Voxel-RCNN~\cite{deng2021voxel} (w/ \texttt{S.A.+C.R.})  & \textbf{90.09} / \textbf{83.10} & \textbf{62.99} / \textbf{58.30} & \textbf{70.20} / \textbf{68.10} & \textbf{59.25} / \textbf{41.51} & \textbf{29.12} / \textbf{23.18} &  \textbf{15.16} / \textbf{13.16}  \\
            \toprule[1pt]
            \toprule[1pt]
            \multirow{2}{*}{only KITTI} & PV-RCNN~\cite{shi2020pv} (w/o \texttt{P.T.})  &  89.41 / 83.15 &  59.09 / 54.73 &  62.25 / 61.71 &  6.58 / 2.54 & 0.22 / 0.16 & 0.03 / 0.01 \\
            & PV-RCNN~\cite{shi2020pv} (w/ \texttt{P.T.} on nuScenes) & 89.26 / 83.14 & \textbf{60.56} / \textbf{55.90} & 63.60 / 62.88 & 13.43 / 5.61 & 0.69 / 0.27 & 0.04 / 0.00  \\
            \toprule[1pt]
            \multirow{2}{*}{only nuScenes} & PV-RCNN~\cite{shi2020pv} (w/o \texttt{P.T.})  & 74.37 / 36.54 &  39.30 / 29.07  &  0.58 / 0.55 & 57.78 / 41.10 & 24.52 / 18.56 & 10.24 / 8.25 \\
            & PV-RCNN~\cite{shi2020pv} (w/ \texttt{P.T} on KITTI) & 69.40 / 38.25 & 33.24 / 24.88 & 1.68 / 1.61  & 53.24 / 36.72  & 20.65 / 17.09  & 8.95 / 7.58  \\
            \toprule[1pt]
            \multirow{4}{*}{KITTI+nuScenes} & PV-RCNN~\cite{shi2020pv} (w/ \texttt{D.M.})  & 87.79 / 77.95 & 55.52 / 48.29 &  59.15 / 55.10 &   41.29 / 21.57  & 10.21 / 7.08 & 1.23 / 1.15 \\
            & PV-RCNN~\cite{shi2020pv} (w/ \texttt{C.A.})  & 88.53 / 77.20 & 47.13 / 39.53 & 44.22 / 41.64 & 46.34 / 25.28 &  12.70 / 9.64 & 2.18 / 1.34  \\
            & PV-RCNN~\cite{shi2020pv} (w/ \texttt{S.A.})  & 87.51 / 78.13 &  56.13 / 49.21 &  61.22 / 58.49  & 56.93 / 40.11 & 20.15 / 15.33 & 10.19 / 8.73  \\
            & PV-RCNN~\cite{shi2020pv} (w/ \texttt{C.R.})  & \textbf{90.93} / 83.56 &  58.96 / 55.78 & 60.92 / 58.13 & 57.76 / 41.31 & 24.65 / 18.96 &  12.19 / 10.13  \\
            & PV-RCNN~\cite{shi2020pv} (w/ \texttt{S.A.+C.R.})  & 89.77 / \textbf{85.49} & 60.03 / 55.58 & \textbf{69.03} / \textbf{66.10} & \textbf{59.08} / \textbf{41.67} & \textbf{25.27} / \textbf{19.26} & \textbf{12.26} / \textbf{10.83}  \\
            \toprule[1pt]
        \end{tabular}
    }
    \end{small}
    \vspace{-0.3cm}
    \caption{Results of joint training on KITTI and nuScenes datasets. The experiment and evaluation settings follow Table~\ref{tab3}.}
    \vspace{-0.35cm}
    \label{tab4}
\end{table*}

\vspace{-0.15cm}
\section{Experiments}
\vspace{-0.1cm}
\subsection{Experimental Setup.}
\vspace{-0.1cm}
\noindent\textbf{Datasets.} We conduct experiments on three commonly-used autonomous driving datasets including Waymo~\cite{sun2020scalability}, nuScenes~\cite{caesar2020nuscenes}, and KITTI~\cite{geiger2012we}. These datasets present: 1) data-level distribution differences caused by different LiDAR types and geographic locations of data collection; and 2) taxonomy-level variations caused by different class annotation definitions. In our experiments, we first consider the task setting of merging two different datasets, and then, perform the study of consolidating all the above datasets.

\noindent\textbf{Implementation Details.} All experiments are implemented using OpenPCDet~\cite{openpcdet2020}. In particular, since we observe that point cloud range differences extremely degrade the cross-dataset detection accuracy as illustrated in Table~\ref{tab2}, we align the point cloud range of all the above datasets to $[-75.2, 75.2]m$ for $X$ and $Y$ axes and $[-2, 4]m$ for $Z$ axis. For all experimental settings, following the common optimization employed by PV-RCNN~\cite{shi2020pv} and Voxel-RCNN~\cite{deng2021voxel}, Adam optimizer with an initial learning rate of $0.01$ is used, and the learning rate decay schedule utilizes the well-known OneCycle strategy. We train the network using a batch-size of 32, a momentum of 0.9 on 8 NVIDIA Tesla A100 GPUs, and the total training epoch is equal to 30. Besides, for the experiments on Waymo-KITTI and nuScenes-KITTI consolidations, the weight decay is set to $0.01$, and for the remaining experiments, the weight decay is set to $0.001$. For Waymo dataset, we only use the uniformly-sampled $20\%$ frames (about 32k frames) for model training.

\begin{table*}[t]
\vspace{-0.3cm}
    \centering
    \begin{small}
    \resizebox{0.90\linewidth}{!}
    {
        \begin{tabular}{c|c|c|c|c|c|c|c}
            \bottomrule[1pt]
            \multirow{2}{*}{Trained on} & \multirow{2}{*}{Baseline Detectors}  & \multicolumn{3}{c|}{Tested on KITTI} & \multicolumn{3}{c}{Tested on Waymo} \\
            \hhline{~~|-|-|-|-|-|-}
            &  & Car & Pedestrian & Cyclist & Vehicle & Pedestrian & Cyclist \\
            \hline
            \multirow{2}{*}{only KITTI} & Voxel-RCNN~\cite{deng2021voxel} (w/o \texttt{P.T.}) & 89.34 / 80.91 & 59.67 / 56.88 &  61.10 / 60.49 &  6.81 / 6.75 & 16.52 / 13.65 & 14.74 / 14.00 \\
            & Voxel-RCNN~\cite{deng2021voxel} (w/ \texttt{P.T.} on Waymo) & 89.51 / 81.41 & 60.30 / 57.10 & 55.53 / 51.34 & 8.70 / 8.62 & 19.14 / 16.01 & 21.87 / 20.83  \\
            \toprule[1pt]
            \multirow{2}{*}{only Waymo} & Voxel-RCNN~\cite{deng2021voxel} (w/o \texttt{P.T.})  &  67.07 / 19.80 & \textbf{65.44} / \textbf{61.92} &  59.48 / 54.10 &  \textbf{75.08} / \textbf{74.60} & \textbf{75.17} / \textbf{68.76} & 65.28 / 64.33 \\
            & Voxel-RCNN~\cite{deng2021voxel} (w/ \texttt{P.T.} on KITTI) & 64.84 / 19.99 & 62.58 / 59.01 & 56.44 / 49.43 & 72.76 / 72.26 & 72.42 / 64.94 & 63.27 / 62.23  \\
            \toprule[1pt]
            \multirow{2}{*}{KITTI+Waymo} & Voxel-RCNN~\cite{deng2021voxel} (w/ \texttt{D.M.})  & 74.53 / 32.11 &  60.11 / 54.85 &  59.69 / 55.94 &   74.35 / 73.85  &  74.80 / 68.39   &   64.87 / 63.95  \\
            & Voxel-RCNN~\cite{deng2021voxel} (w/ \texttt{S.A.+C.R.})  &  \textbf{90.03} / \textbf{82.39} & 62.51 / 57.01 & \textbf{69.52} / \textbf{66.30} & 74.83 / 74.33 &  74.79 / 68.24 &  \textbf{66.83} / \textbf{65.82}  \\
            \toprule[1pt]
            \toprule[1pt]
            \multirow{2}{*}{only KITTI} & PV-RCNN~\cite{shi2020pv} (w/o \texttt{P.T.})  &  89.41 / 83.15 &  59.09 / 54.73 &  62.25 / 61.71 &  2.98 / 2.94  & 7.99 / 6.56 & 5.84 / 5.54 \\
            & PV-RCNN~\cite{shi2020pv} (w/ \texttt{P.T.} on Waymo) & 89.40 / \textbf{83.42} & 62.69 / 58.86 & 59.96 / 59.43 & 8.75 / 8.64 & 12.12 / 9.90  & 9.20 / 8.76  \\
            \toprule[1pt]
            \multirow{2}{*}{only Waymo} & PV-RCNN~\cite{shi2020pv} (w/o \texttt{P.T.}) & 56.20 / 54.81 &  60.04 / 57.06  &  54.29 / 50.05 & 74.97 / 74.46
             & \textbf{73.41} / \textbf{66.57} & \textbf{64.58} / \textbf{63.49} \\
            & PV-RCNN~\cite{shi2020pv} (w/ \texttt{P.T.} on KITTI) & 69.25 / 25.91 & 59.16 / 55.92 & 56.09 / 50.50 & 71.08 / 70.54  &  70.12 / 62.91  &  62.37 / 61.40  \\
            \toprule[1pt]
            \multirow{2}{*}{KITTI+Waymo} & PV-RCNN~\cite{shi2020pv} (w/ \texttt{D.M.})  & 87.49 / 68.35 & \textbf{62.84} / \textbf{60.06} &  68.09 / 65.75 &   50.68 / 50.31  & 58.76 / 52.59 & 55.14 / 54.17  \\
            & PV-RCNN~\cite{shi2020pv} (w/ \texttt{S.A.+C.R.})  & \textbf{89.42} / 83.15 & 60.85 / 57.49 & \textbf{71.61} / \textbf{65.88} & \textbf{75.07} / \textbf{74.54} & 72.95 / 66.08 & 63.80 / 62.92  \\
            \toprule[1pt]
        \end{tabular}
    }
    \end{small}
    \vspace{-0.3cm}
    \caption{Results of joint training on KITTI and Waymo. The experiment and evaluation settings follow Table~\ref{tab3}.}
    \vspace{-0.3cm}
    \label{tab5}
\end{table*}

\begin{table}[]
\small
\centering
\setlength{\tabcolsep}{1.8mm}{
\resizebox{\linewidth}{!}
{
\begin{tabular}{ccccc}
\hline
Trained on & Tested on K & Tested on N & Tested on W & Avg. on KNW \\
\hline
K & 89.34 / 80.91 & 11.37 / 4.64 & 6.81 / 6.75  & 35.84 / 30.77 \\
N & 69.41 / 33.48 & 53.63 / 39.05 & 36.77 / 36.50  & 53.27 / 36.34 \\
W & 67.07 / 19.80 & 34.10 / 17.31  & 75.08 / 74.60  & 58.75 / 37.23 \\
K+N+W (Uni3D) & \textbf{89.65} / \textbf{83.41} & \textbf{60.42} / \textbf{42.30} &  \textbf{75.47} / \textbf{74.97}  & \textbf{75.18} / \textbf{66.89} \\
\hline
\end{tabular}}
}
\vspace{-0.3cm}
\caption{Results for car class of jointly train on K (denoting KITTI), N (denoting nuScenes), and W (denoting Waymo) using Voxel-RCNN~\cite{deng2021voxel}, and Avg. denotes the average detection accuracy evaluated on all the three datasets.}
\label{tab6}
\vspace{-0.45cm}
\end{table}

\begin{table*}[h]
\vspace{-0.3cm}
    \centering
    \begin{small}
    \resizebox{0.91\linewidth}{!}
    {
        \begin{tabular}{c|c|c|c|c|c|c|c|c}
            \bottomrule[1pt]
            \multirow{2}{*}{Trained on} & \multirow{2}{*}{Baseline Detectors} & \multirow{2}{*}{\#nuScenes} & \multicolumn{3}{c|}{Tested on KITTI} & \multicolumn{3}{c}{Tested on nuScenes} \\
            \hhline{~~~|-|-|-|-|-|-}
            &  &  & Car & Pedestrian & Cyclist & Car & Pedestrian & Cyclist \\
            \hline
            \multirow{1}{*}{only nuScenes} & Voxel-RCNN~\cite{deng2021voxel} & $100\%$ & - & - & - &  53.63 / 39.05 & 22.47 / 17.85 & 10.86 / 9.70 \\
            \multirow{1}{*}{only nuScenes} & Voxel-RCNN~\cite{deng2021voxel} & $10\%$ & - & - & - &  45.42 / 31.09  & 10.39 / 7.16  & 1.55 / 0.89  \\
            \multirow{1}{*}{only nuScenes} & Voxel-RCNN~\cite{deng2021voxel} & $5\%$ & - & - & - &  30.01 / 16.15  & 4.70 / 2.56  & 0.06 / 0.05  \\
            \multirow{1}{*}{only nuScenes} & Voxel-RCNN~\cite{deng2021voxel} & $1\%$ & - & - & - &  0.00 / 0.00  & 0.00 / 0.00  & 0.00 / 0.00  \\
            \toprule[1pt]
            \multirow{1}{*}{KITTI+nuScenes} 
            & Voxel-RCNN~\cite{deng2021voxel} (ours)  &  $100\%$  & 90.09 / 83.10 & 62.99 / 58.30 & 70.20 / 68.10 & 59.25 / 41.51  &  29.12 / 23.18  & 15.16 / 13.16 \\
            & Voxel-RCNN~\cite{deng2021voxel} (ours)  &  $10\%$  & 88.81 / 81.75 & 60.09 / 56.61 & 70.03 / 68.54 & 52.08 / 34.40  &  20.40 / 15.60  & 8.42 / 7.40 \\
            & Voxel-RCNN~\cite{deng2021voxel} (ours)  &  $5\%$  & 89.10 / 81.86 & 59.17 / 54.42 & 73.30 / 70.25 & 51.81 / 34.43  &  19.82 / 14.94  & 5.52/ 4.58 \\
            & Voxel-RCNN~\cite{deng2021voxel} (ours)  &  $1\%$  & 89.06 / 81.55 & 56.74 / 52.28 & 71.11 / 69.06 & 44.74 / 28.28  &  15.94 / 11.11  & 1.28 / 0.99 \\
            \toprule[1pt]
            \toprule[1pt]
            \multirow{1}{*}{only nuScenes} & PV-RCNN~\cite{shi2020pv} & $100\%$ & - & - & - &  57.78 / 41.10 & 24.52 / 18.56 & 10.24 / 8.25 \\
            \multirow{1}{*}{only nuScenes} & PV-RCNN~\cite{shi2020pv} & $10\%$ & - & - & - &  50.39 / 31.68  &  13.64 / 8.75   &  0.85 / 0.51  \\
            \multirow{1}{*}{only nuScenes} & PV-RCNN~\cite{shi2020pv} & $5\%$ & - & - & - &   35.87 / 19.76  &  5.89 / 3.15   &  0.00 / 0.00  \\
            \multirow{1}{*}{only nuScenes} & PV-RCNN~\cite{shi2020pv} & $1\%$ & - & - & - &   0.08 / 0.01  &  0.02 / 0.01   &  0.00 / 0.00  \\
            \toprule[1pt]
            \multirow{1}{*}{KITTI+nuScenes} 
            & PV-RCNN~\cite{shi2020pv} (ours)  &  $100\%$  & 89.77 / 85.49 & 60.03 / 55.58 & 69.03 / 66.10 & 59.08 / 41.67  &  25.27 / 19.26  & 12.26 / 10.83 \\
            & PV-RCNN~\cite{shi2020pv} (ours)  &  $10\%$  & 88.99 / 83.12 & 57.06 / 52.48 & 71.14 / 70.60 & 51.75 / 33.85  & 15.60 / 10.78  & 3.33 / 2.09 \\
            & PV-RCNN~\cite{shi2020pv} (ours)  &  $5\%$  &  88.95 / 82.83  & 56.62 / 53.25 & 71.99 / 69.86 & 50.32 / 34.35  &  16.11 / 11.20  & 2.59 / 2.00 \\
            & PV-RCNN~\cite{shi2020pv} (ours)  &  $1\%$  & 88.92 / 82.81  & 55.22 / 51.84 & 71.12 / 69.73 &  41.09 / 25.38  &  11.27 / 7.00  & 0.60 / 0.33 \\
            \toprule[1pt]
        \end{tabular}
    }
    \end{small}
    \vspace{-0.3cm}
    \caption{Results of reducing the number of samples in nuScenes dataset under the nuScenes-KITTI consolidation setting.} 
    \vspace{-0.40cm}
    \label{tab7}
\end{table*}

\vspace{-0.25cm}
\subsection{Design of Comparison Baselines}
\vspace{-0.15cm}
\noindent1) w/o \texttt{P.T.(Single-dataset)}: We employ the off-the-shelf 3D detectors, \textit{e.g.}, Voxel-RCNN~\cite{deng2021voxel} and PV-RCNN~\cite{shi2020pv}, as the baseline detection model, which is trained \underline{from scratch} and evaluated within a single dataset.

\noindent2) \texttt{P.T.(Pre-training)}: Since the MDF setting allows the detector to access the annotated data from both datasets, we first pre-train the baseline detector on another dataset, and fine-tune the detector on the current dataset.

\noindent3) \texttt{D.M.(Direct Merging)}: By simply combing multiple 3D datasets into a merged dataset, a single-dataset baseline detector is able to train from the merged dataset using a common detection loss, which can be regarded as a direct method to verify whether the existing 3d models can be improved under the directly-merged datasets. \textbf{Note that} for such baseline, we align the point cloud range and merge the label space of each dataset for multi-dataset training.

\vspace{-0.02cm}
\noindent4) \texttt{C.A.}: Coordinate-origin Alignment baseline is designed to alleviate the sensor installation position differences. As previously described, to train detectors from multiple 3D datasets with inconsistent point cloud ranges, we \textbf{have to} align the point-cloud-range of all datasets. But such a point-range-level alignment operation would cause the distribution variations in coordinate origin and center point of objects. To this end, following previous 3D cross-dataset research~\cite{yang2021st3d,yang2022st3d++,wei2022lidar}, the coordinate origin of different datasets needs to be transferred to the ground plane. For KITTI and nuScenes datasets, the coordinate shift is set to $1.6$m and $1.8$m along the height direction. We use the \texttt{C.A.} baseline to check the effectiveness of the existing origin alignment method on multi-dataset 3D object detection.

\noindent5) \texttt{S.A.}: As introduced in Sec.~\ref{subsec:method}, Statistics-level Alignment baseline aims to reduce the statistical distribution (\textit{e.g.}, $\mu^j$ and $\sigma^j$) differences of the learned features between datasets.

\noindent6) \texttt{C.R.}: Coupling-and-Recoupling baseline tries to mine reusable dataset-agnostic representations across multiple datasets, and outputs the dataset-specific semantic representations for better single-dataset detection accuracy.

\noindent\textbf{Evaluation Metric.} We adopt the officially released evaluation tools for evaluating our all baselines, where for KITTI and nuScenes datasets, the AP in both the Bird's Eye View (BEV) and 3D over 40 recall positions are reported, and for Waymo dataset, we employ the Average Precision (AP) and Average Precision re-weighted by Heading (APH) of each class for model evaluation. We report the moderate case results for KITTI dataset, and LEVEL\_1 metric on Waymo dataset, and please refer to Appendix for the results using LEVEL\_2 metric. AP is evaluated under an IoU threshold of 0.7 for car category (Vehicle on Waymo) and 0.5 for pedestrian and cyclist classes. In this paper, all experimental results are reported on the official validation set.

\vspace{-0.10cm}
\subsection{Results of Multi-Dataset 3D Object Detection}
\label{sec4.3}
\vspace{-0.10cm}
\noindent\textbf{Results on Waymo-nuScenes, nuScenes-KITTI, Waymo-KITTI Consolidations.} To investigate the feasibility of training 3D baseline detectors from multiple public datasets, we conduct experiments by selecting two representative 3D datasets from three widely-used autonomous driving datasets: Waymo~\cite{sun2020scalability}, KITTI~\cite{geiger2012we}, and nuScenes~\cite{caesar2020nuscenes}. According to the results from Table~\ref{tab3} to~\ref{tab5}, we can observe the following five important findings: 

\noindent\textit{1) Large gap between 3D datasets}: We first train the baseline detector only on a single dataset (\textit{e.g.}, Waymo or nuScenes), and evaluate this well-trained baseline on two different datasets (\textit{e.g.}, Waymo and nuScenes). As can be seen in Table~\ref{tab3}, the baseline performs well only on its original training dataset (\textit{e.g.}, $75.08 \%$ AP for Vehicle). When the baseline detector is deployed to nuScenes dataset, its detection accuracy is seriously degraded (only $34.34 \%$ AP for Car). This is mainly because the single-dataset detection model is overfitted to its training dataset, yet fails to consider the source-to-target dataset shift. The same accuracy drop issue can also be observed on another baseline detector such as PV-RCNN~\cite{shi2020pv}.

\noindent\textit{2) Pre-trained model cannot work well under MDF setting}: Another way that simultaneously improves the detection accuracy of the baseline detector for Waymo and nuScenes is fully-supervised pre-training. Such a way means that we first pre-train the baseline on the fully-labeled nuScenes (or Waymo), and fine-tune the well-trained model on Waymo (or nuScenes). By comparing \texttt{P.T} and w/o~\texttt{P.T.} baselines, we observe from Table~\ref{tab3} that, although the model has been pre-trained on nuScenes, the detection accuracy for nuScenes is still unsatisfactory, which is due to that the model has been fine-tuned to Waymo, forgetting the knowledge learned from the previous pre-trained dataset. 

\noindent\textit{3) 3D single-dataset training paradigm cannot work well under MDF setting}: By comparing w/o \texttt{P.T.} and \texttt{D.M} baselines from Table~\ref{tab3}, it can be observed that the typical 3D detectors (\textit{e.g.}, Voxel-RCNN~\cite{deng2021voxel} and PV-RCNN~\cite{shi2020pv}) trained on the merged dataset cannot achieve a high detection accuracy on both datasets. 

\noindent\textit{4) Coordinate-origin alignment boosts the MDF accuracy}: By comparing \texttt{C.A.} and \texttt{D.M.} baselines, it can be concluded that coordinate-origin shift scheme can reduce the negative impact caused by the point-cloud-range alignment operation. Besides, it should be noted that for KITTI-nuScenes consolidation setting, applying the coordinate-origin shift operation for raw point clouds yields a severe detection accuracy drop of the Pedestrian and Cyclist classes in KITTI datasets. This may be due to that the preset coordinate-origin shift parameters are shared across different classes, which is sensitive to correct the distributions for the classes with few-shot samples, such as Pedestrian and Cyclist. Thus, sharing the same parameter of coordinate-origin shift between classes is harmful to some scenarios, and needs to be further studied in our future work.

\noindent\textit{5) The effectiveness and generality of each designed module}: As reported in Tables~\ref{tab3},~\ref{tab4}, and~\ref{tab5}, the average detection results achieved by Uni3D exceed those of all designed baselines, verifying the effectiveness of Uni3D in learning from multiple 3D datasets. Furthermore, we also conduct experiments by selecting PV-RCNN~\cite{shi2020pv} as another baseline detector and repeat the above experiments, observing consistent detection accuracy gains.

\noindent\textbf{Results on Waymo-KITTI-nuScenes Consolidation.} Table~\ref{tab6} shows the results of jointly training the Voxel-RCNN~\cite{deng2021voxel} from Waymo, nuScenes, and KITTI. Also, Uni3D achieves high detection results simultaneously on multiple datasets.

\vspace{-0.15cm}
\subsection{Further Analyses}
\label{analyses}
\vspace{-0.10cm}

\noindent\textbf{Uni3D: Reduce the Data Acquisition Cost.} Considering a real-world scenario: we may not be able to collect massive LiDAR data for a new scene due to the expensive data acquisition cost. Uni3D gives another option for addressing such a dilemma, namely, training on the combined set between the few-shot data from the new scene and the full data from the previous well-constructed dataset. It can be seen from Table~\ref{tab7} that the detection accuracy on nuScenes dataset achieved by our Uni3D outperforms the one only using few-shot nuScenes itself. The improvement mainly comes from the ability of Uni3D to learn more generalizable features, which is less prone to over-fitting under few-shot samples, further verifying the superiority of our Uni3D in reducing data dependency.

\noindent\textbf{Uni3D: Strengthen the Zero-shot and Domain Adaptation Ability.} Another advantage of Uni3D is that it can largely boost the zero-shot learning ability of the baseline detector, by learning generalizable features from multiple datasets. As shown in Table~\ref{tab8}, by utilizing the pre-trained model (including 3D and 2D backbones) provided by Uni3D, the zero-shot inference accuracy is significantly improved (from $68.15\% AP_{BEV}$ to $73.51\% AP_{BEV}$). One possible reason for such improvement is that Uni3D learns about the potential inter-dataset variations by jointly training on Waymo and nuScenes, and the Waymo-to-nuScenes dataset variations are beneficial to recognize an unforeseen domain. Further, the pre-trained parameters generated by our Uni3D also can further enhance the domain adaptability of the existing UDA models such as ST3D~\cite{yang2021st3d}.

\begin{table}[]
\small
\centering
\setlength{\tabcolsep}{1.8mm}{
\resizebox{0.97\linewidth}{!}
{
\begin{tabular}{cccc}
\hline
\multirow{2}{*}{Methods} & \multirow{2}{*}{Baseline Models} & \multirow{2}{*}{Pre-trained on} & Tested on KITTI \\
 &   &  &  $\text{AP}_{\text{BEV}}$ / $\text{AP}_{\text{3D}}$  \\ 
\hline
Source-only & PV-RCNN & Waymo & 61.18 / 22.01 \\
Source-only & PV-RCNN & nuScenes & 68.15 / 37.17 \\
Source-only & PV-RCNN (ours) & \textbf{Waymo+nuScenes} & \textbf{73.51} / \textbf{39.71} \\
ST3D~\cite{yang2021st3d} (w/ \texttt{SN}) & PV-RCNN & Waymo & 86.65 / 76.86 \\
ST3D~\cite{yang2021st3d} (w/ \texttt{SN}) & PV-RCNN & nuScenes & 84.29 / 72.94 \\
ST3D~\cite{yang2021st3d} (w/ \texttt{SN}) & PV-RCNN (ours) & \textbf{Waymo+nuScenes} & \textbf{88.25} / \textbf{77.01}  \\
\hline
\end{tabular}}
}
\vspace{-0.3cm}
\caption{Generalization study from two aspects including: 1) Zero-shot detection accuracy on KITTI, and 2) Model adaptability coupled with the off-the-shelf UDA method (ST3D~\cite{yang2021st3d}). Source-only denotes that the model is trained on the source domain and directly tested on the target domain.}
\label{tab8}
\vspace{-0.45cm}
\end{table}

\vspace{-0.10cm}
\section{Conclusion}
\vspace{-0.10cm}
In this work, for the first time, we study how to train a unified 3D detection model using the off-the-shelf public 3D benchmarks, and present a unified 3D detection framework (Uni3D) consisting of a data-level correction operation and a semantic-level feature coupling-and-recoupling module, which can be easily combined with the existing 3D detectors. We conduct extensive experiments on many public benchmarks, and the results show the effectiveness of Uni3D in obtaining dataset-level generalizable features. 

\section*{Acknowledgement}
This work is supported by Science and Technology Commission of Shanghai Municipality (grant No. 22DZ1100102).

{\small
\bibliographystyle{ieee_fullname}
\bibliography{egbib}
}

\clearpage
\appendix

\section{Appendix}

\subsection{Uni3D Implementation}
\label{sec:details}

\noindent \textbf{Detailed Structure of the Designed SE Module.}
As mentioned in Sec.~3.3 in our main text, the Squeeze-and-Excitation (SE) Network is utilized to obtain the re-scaled dataset-specific features. The visualization network of the designed SE network is shown in Fig.~\ref{fig:se}. 

\setcounter{figure}{4}
\begin{figure}[h]
\centering
\includegraphics[width=7.0cm]{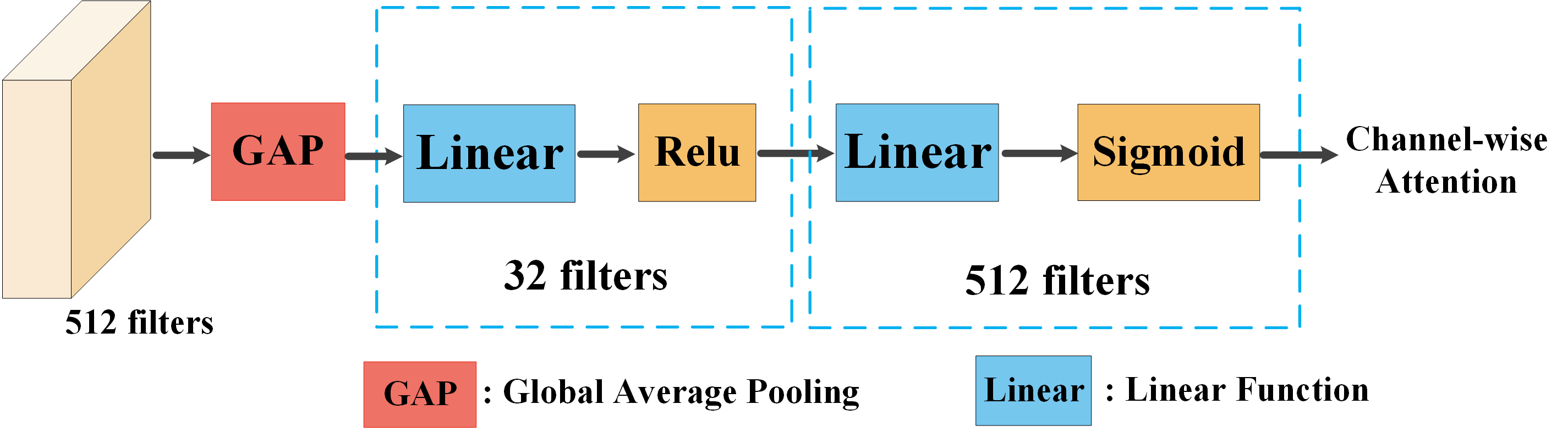}
\vspace{-0.15cm}
\caption{Overview of the designed Squeeze-and-Excitation (SE) Network.}
\label{fig:se}
\end{figure}

\setcounter{figure}{5}
\begin{figure*}
\centering
\includegraphics[width=10.9cm]{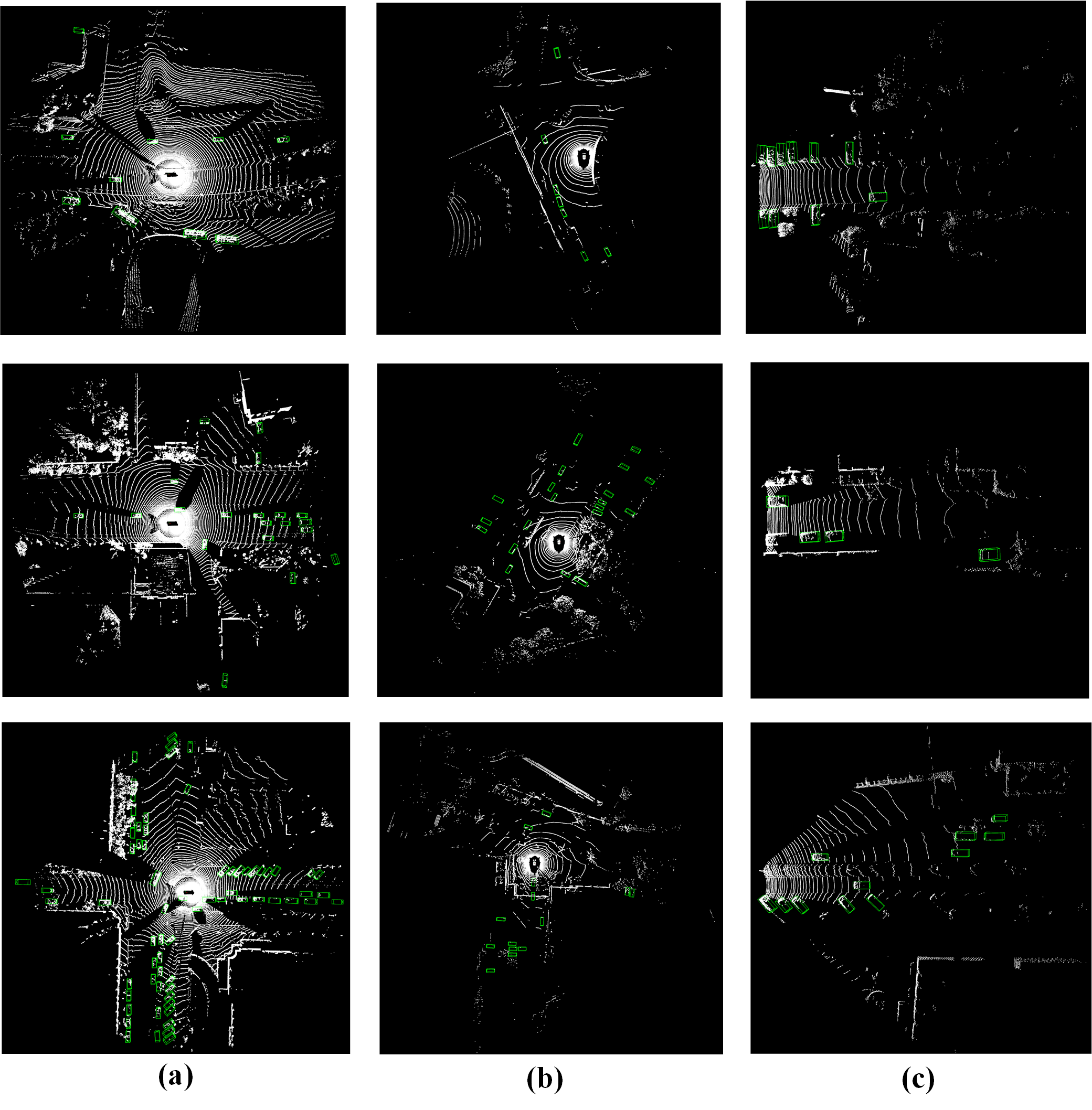}
\vspace{-0.5cm}
\caption{Comparisons across different 3D autonomous driving datasets including: (a) Waymo, (b) nuScenes, and (c) KITTI datasets, where the \textcolor{green}{green boxes} denote the 3D bounding boxes. We can observe that different 3D datasets present significant differences in LiDAR point cloud range, distribution of foreground object, and \textit{etc}. }
\label{fig:compare}
\end{figure*}

\setcounter{table}{8}
\begin{table*}[t]
\small
\centering
\setlength{\tabcolsep}{1.8mm}{
\resizebox{0.90\linewidth}{!}
{
\begin{tabular}{ccc|cccc}
\hline
\multirow{2}{*}{Methods} & \multirow{2}{*}{Waymo Range} & \multirow{2}{*}{KITTI Range} & \multicolumn{2}{c}{Tested on Waymo} & \multicolumn{2}{c}{Tested on KITTI} \\
\hhline{~~~|----}
 &  &  &  Pedestrian & Cyclist  & Pedestrian & Cyclist  \\ 
\hline
\multirow{3}{*}{Not Align.} & L=[-75.2, 75.2]m & L=[0.0, 70.4]m & \multirow{3}{*}{26.61 / 23.37} & \multirow{3}{*}{26.47 / 25.87} & \multirow{3}{*}{59.17 / 57.16} & \multirow{3}{*}{62.16 / 61.25}  \\
& W=[-75.2, 75.2]m & W=[-40.0, 40.0]m  & \\
& H=[-2.0, 4.0]m & H=[-3.0, 1.0]m & \\
\toprule[0.5pt]
\multirow{3}{*}{Align. (w/ ours)} & L=[-75.2, 75.2]m & L=[-75.2, 75.2]m & \multirow{3}{*}{\textbf{74.79} / \textbf{68.24}} & 
 \multirow{3}{*}{\textbf{66.83} / \textbf{65.82}} & \multirow{3}{*}{\textbf{62.51} / \textbf{57.01}}  & \multirow{3}{*}{\textbf{69.52} / \textbf{66.30}}  \\
& W=[-75.2, 75.2]m & W=[-75.2, 75.2]m  & \\
& H=[-2.0, 4.0]m & H=[-2.0, 4.0]m & \\
\toprule[0.5pt]
\toprule[0.5pt]
\multirow{2}{*}{Methods} & \multirow{2}{*}{nuScenes Range} & \multirow{2}{*}{KITTI Range} & \multicolumn{2}{c}{tested on nuScenes} & \multicolumn{2}{c}{tested on KITTI} \\
\hhline{~~~|----}
 &  &  &  Pedestrian & Cyclist  & Pedestrian & Cyclist  \\ 
\hline
\multirow{3}{*}{Not Align.} & L=[-51.2, 51.2]m & L=[0.0, 70.4]m & \multirow{3}{*}{11.34 / 9.08} & \multirow{3}{*}{8.67 / 6.37} & \multirow{3}{*}{60.05 / 54.43} & \multirow{3}{*}{63.98 / 63.06} \\
& W=[-51.2, 51.2]m & W=[-40.0, 40.0]m  & \\
& H=[-5.0, 3.0]m & H=[-3.0, 1.0]m & \\
\toprule[0.5pt]
\multirow{3}{*}{Align. (w/ ours)} & L=[-75.2, 75.2]m & L=[-75.2, 75.2]m & \multirow{3}{*}{\textbf{29.12} / \textbf{23.18}} & \multirow{3}{*}{\textbf{15.16} / \textbf{13.16}} & \multirow{3}{*}{\textbf{62.99} / \textbf{58.30}} & \multirow{3}{*}{\textbf{70.20} / \textbf{68.10}} \\
& W=[-75.2, 75.2]m & W=[-75.2, 75.2]m  & \\
& H=[-2.0, 4.0]m & H=[-2.0, 4.0]m & \\
\hline
\end{tabular}}
}
\vspace{-0.3cm}
\caption{Inconsistent LiDAR ranges will cause multi-dataset detection accuracy drop, where we show the results in Pedestrian and Cyclist categories with or without point-range alignment.}
\label{tab:range}
\end{table*}

\begin{table*}[htbp]
    \centering
    \begin{small}
    \resizebox{0.94\linewidth}{!}
    {
        \begin{tabular}{c|c|c|c|c|c|c|c}
            \bottomrule[1pt]
            \multirow{2}{*}{Model} & \multirow{2}{*}{Option}  & \multicolumn{3}{c|}{Tested on Waymo} & \multicolumn{3}{c}{Tested on nuScenes} \\
            \hhline{~~|-|-|-|-|-|-}
            &  & Vehicle & Pedestrian & Cyclist & Car & Pedestrian & Cyclist \\
            \hline
            Voxel-RCNN (Direct Merging) & -  & 66.67 / 66.23 & 60.36 / 54.08 &  52.03 / 51.25 &  51.40 / 31.68   & 15.04 / 9.99 &  5.40 / 3.87 \\
            Voxel-RCNN (w/ \texttt{C.A.+S.A.+C.R.}) & \textbf{BEV feature copy} & \textbf{75.26} / \textbf{74.77} & \textbf{75.46} / \textbf{68.75} & \textbf{65.02} / \textbf{63.12}  & 60.18 / \textbf{42.23}   &  \textbf{30.08} / \textbf{24.37}   &  \textbf{14.60} / 12.32 \\
            Voxel-RCNN (w/ \texttt{C.A.+S.A.+C.R.}) & \textbf{BEV feature mask} & 73.78 / 73.29  &  72.67 / 66.32  &   64.20 / 62.81  & \textbf{60.37} / 40.66  &  29.57 / 23.51  &  14.13 / \textbf{12.42}  \\
            \bottomrule[1pt]
        \end{tabular}
    }
    \end{small}
    \vspace{-0.3cm}
    \caption{Options for inference usage of the \texttt{C.R.} module: The model is jointly trained on Waymo and nuScenes, and evaluated on the validation of Waymo and nuScenes.}
    \label{tab:inference}
\end{table*}

\begin{table*}[h]
\centering
\begin{small}
\resizebox{0.9\linewidth}{!}{
    \begin{tabular}{ccccc}
    \hline
    Model              & Ensemble?   & Vehicle      & Pedestrian       & Cyclist         \\ \hline
    Voxel-RCNN (w/ \texttt{C.A.+S.A.})  & No      & 75.16 / 74.67     & 74.83 / 68.07     & 64.68 / \textbf{63.73}          \\
    Voxel-RCNN (w/ \texttt{C.A.+S.A.+E.N.})  & Yes      & 70.59 / 70.07     & 73.56 / 66.86   &  62.75 / 61.82       \\
    Voxel-RCNN (w/ \texttt{C.A.+S.A.+C.R.})  & No   & \textbf{75.26} / \textbf{74.77}     & \textbf{75.46} / \textbf{68.75}     & \textbf{65.02} / 63.12             \\ \hline
    \end{tabular}
}
\end{small}
\vspace{-0.3cm}
\caption{Comparisons against model ensemble: The model is trained on Waymo and nuScenes, and evaluated on the validation of Waymo. \texttt{E.N.} is the dataset-level model ensemble.}
\label{tab:ensembling}
\end{table*}

\noindent \textbf{Dataset Introduction.}
Waymo Open Dataset~\cite{sun2020scalability} covers 1000 fully-annotated sequences which are collected in USA using a 64-beam LiDAR. Waymo dataset is divided into a train set with 798 sequences (158081 samples) and a validation set with 202 sequences (39987 samples). For all Waymo-related dataset consolidation experiments, we only use $20\%$ frames (About 32k frames) of all the training samples for saving the model training time.

nuScenes~\cite{caesar2020nuscenes} dataset is constructed using 32-beam LiDAR, where all data are acquired from different countries and regions, such as Singapore and Boston, USA. nuScenes includes 28130 training frames and 6019 validation frames. For our experiments, we use the 28130 training frames for the model training, and evaluate our all models on the 6019 validation frames.

KITTI dataset~\cite{geiger2012we} is one of the most popular datasets in the autonomous driving community, consisting of 7481 training frames, where we divide the 7481 training frames into a train set (including 3712 samples) and a validation set (including 3769 samples). This data splitting method is consistent with that of previous works~\cite{yang2021st3d,yang2022st3d++,wei2022lidar}. Similar to Waymo dataset, point clouds in KITTI dataset are collected using a 64-beam LiDAR. Note that since KITTI only provides the annotations from the Front-camera Of View (FOV), we remove the point clouds and prediction results outside of the FOV during the model evaluation phase.

\subsection{Challenges of Merging Datasets}
\label{merge-challenge}

With the rapid increase of autonomous driving datasets, our study focuses on how to train a unified 3D detector from such continuously increasing 3D datasets.

\begin{table*}[h]
\centering
\begin{small}
\resizebox{0.75\linewidth}{!}{
    \begin{tabular}{c|c|ccc}
    \hline
    Model   &  Module               & Vehicle      & Pedestrian       & Cyclist         \\ \hline
    Voxel-RCNN & \texttt{C.A.+S.A.+C.R.} w/o AT  & 75.06 / 74.56     & 74.90 / 68.34     & 64.31 / 63.36          \\
    Voxel-RCNN  & \texttt{C.A.+S.A.+C.R.} w/o SE     & 74.37 / 73.85     & 74.42 / 67.77    & 64.47 / \textbf{63.50}          \\
    Voxel-RCNN  & \texttt{C.A.+S.A.+C.R.}   & \textbf{75.26} / \textbf{74.77}     & \textbf{75.46} / \textbf{68.75}     & \textbf{65.02} / 63.12             \\ \midrule\
    PV-RCNN & \texttt{C.A.+S.A.+C.R.} w/o AT  & 74.25 / 73.71    & 70.85 / 63.89     & 62.03 / 60.95          \\
    PV-RCNN  & \texttt{C.A.+S.A.+C.R.} w/o SE     & 74.79 / 74.24     & 73.69 / 66.65     & 62.68 / 61.04          \\
    PV-RCNN  & \texttt{C.A.+S.A.+C.R.}   & \textbf{75.54} / \textbf{74.90}    & \textbf{74.12} / \textbf{66.90}     & \textbf{63.28} / \textbf{62.12}             \\ \hline
    \end{tabular}
}
\end{small}
\vspace{-0.3cm}
\caption{Ablation studies within the \texttt{C.R.} module: AT and SE denote the Attention Design and Squeeze-and-Excitation.}
\label{tab:at_se}
\end{table*}

\begin{table*}[t]
    \centering
    \begin{small}
    \resizebox{0.80\linewidth}{!}
    {
        \begin{tabular}{c|c|c|c|c}
            \bottomrule[1pt]
            \multirow{2}{*}{Trained on} & \multirow{2}{*}{Baseline Detectors}  & \multicolumn{3}{c}{Avg. on Waymo+nuScenes} \\
            \hhline{~~|-|-|-}
            &  & Vehicle\&Car & Pedestrian & Cyclist \\
            \hline
            \multirow{2}{*}{only Waymo} & Voxel-RCNN~\cite{deng2021voxel} (w/o \texttt{P.T.})  & 46.20  &  38.43  &  32.64 \\
            & Voxel-RCNN~\cite{deng2021voxel} (w/ \texttt{P.T.} on nuScenes) & 48.71  & 38.08 &  32.97  \\
            \toprule[1pt]
            \multirow{2}{*}{only nuScenes} & Voxel-RCNN~\cite{deng2021voxel} (w/o \texttt{P.T.})   &  37.91  & 11.25  &  6.10  \\
            & Voxel-RCNN~\cite{deng2021voxel} (w/ \texttt{P.T.} on Waymo) & 22.63 & 8.62 & 2.91  \\
            \toprule[1pt]
            \multirow{4}{*}{Waymo+nuScenes} & Voxel-RCNN~\cite{deng2021voxel} (w/ \texttt{D.M.})  & 49.18 & 35.18 & 27.95  \\
            & Voxel-RCNN~\cite{deng2021voxel} (w/ \texttt{C.A.}) & 49.22 & 37.20 &  27.98 \\
            & Voxel-RCNN~\cite{deng2021voxel} (w/ \texttt{C.A.+S.A.})  & 58.00 & 47.91 & 36.17  \\
            & Voxel-RCNN~\cite{deng2021voxel} (w/ \texttt{C.A.+C.R.})  & 58.41 & 49.03 & 37.94  \\
            & Voxel-RCNN~\cite{deng2021voxel} (w/ \texttt{C.A.+S.A.+C.R.})  & \textbf{58.75} & \textbf{49.92} & \textbf{38.67}  \\
            \toprule[1pt]
            \toprule[1pt]
            \multirow{2}{*}{only Waymo} & PV-RCNN~\cite{shi2020pv} (w/o \texttt{P.T.})   & 46.26 & 37.68 & 32.30  \\
            & PV-RCNN~\cite{shi2020pv} (w/ \texttt{P.T.} on nuScenes) & 46.12  & 37.43  &  32.04   \\
            \toprule[1pt]
            \multirow{2}{*}{only nuScenes} & PV-RCNN~\cite{shi2020pv} (w/o \texttt{P.T.})  & 41.06  & 11.57 &  4.62  \\
            & PV-RCNN~\cite{shi2020pv} (w/ \texttt{P.T.} on Waymo) & 43.06  & 12.49  &  8.98  \\
            \toprule[1pt]
            \multirow{4}{*}{Waymo+nuScenes} & PV-RCNN~\cite{shi2020pv} (w/ \texttt{D.M.})  & 48.33 & 31.77 &  28.77  \\
            & PV-RCNN~\cite{shi2020pv} (w/ \texttt{C.A.})  & 49.06  &  33.36 &  28.54 \\
            & PV-RCNN~\cite{shi2020pv} (w/ \texttt{C.A.+S.A})  & 58.15 &  44.16  &  35.27   \\
            & PV-RCNN~\cite{shi2020pv} (w/ \texttt{C.A.+C.R.})  & 58.02  &  46.94  & 35.22  \\
            & PV-RCNN~\cite{shi2020pv} (w/ \texttt{C.A.+S.A.+C.R.})  & \textbf{59.10} & \textbf{47.99} & \textbf{37.58}  \\
            \toprule[1pt]
        \end{tabular}
    }
    \end{small}
    \vspace{-0.3cm}
    \caption{Average (Avg.) detection results of joint training on Waymo and nuScenes datasets. Here, we report the car (Vehicle on Waymo), pedestrian, and cyclist results under IoU threshold of 0.7, 0.5, and 0.5, respectively, and utilize $\text{AP}$ of LEVEL 1 metric on Waymo, and $\text{AP}_{\text{3D}}$ over 40 recall positions on nuScenes. The best detection results are marked using \textbf{bold}.}
    \label{tab:avg_waymo_nusc}
\end{table*}

\begin{table*}[h]
    \centering
    \begin{small}
    \resizebox{0.80\linewidth}{!}
    {
        \begin{tabular}{c|c|c|c|c}
            \bottomrule[1pt]
            \multirow{2}{*}{Trained on} & \multirow{2}{*}{Baseline Detectors}  & \multicolumn{3}{c}{Avg. on KITTI+nuScenes}  \\
            \hhline{~~|-|-|-}
            &  & Car & Pedestrian & Cyclist \\
            \hline
            \multirow{2}{*}{only KITTI} & Voxel-RCNN~\cite{deng2021voxel} (w/o \texttt{P.T.})  & 42.78  &  28.50  & 30.25 \\
            & Voxel-RCNN~\cite{deng2021voxel} (w/ \texttt{P.T.} on nuScenes) & 43.39 & 28.18 & 27.09 \\
            \toprule[1pt]
            \multirow{2}{*}{only nuScenes} & Voxel-RCNN~\cite{deng2021voxel} (w/o \texttt{P.T.})  & 36.27  &  18.53  &  5.07 \\
            & Voxel-RCNN~\cite{deng2021voxel} (w/ \texttt{P.T.} on KITTI) &  40.15  & 25.58  & 8.42  \\
            \toprule[1pt]
            \multirow{4}{*}{KITTI+nuScenes} & Voxel-RCNN~\cite{deng2021voxel} (w/ \texttt{D.M.})  &  47.1  &  31.44  &  30.45  \\
            & Voxel-RCNN~\cite{deng2021voxel} (w/ \texttt{C.A.})  &  52.60  &  32.80  & 22.77  \\
            & Voxel-RCNN~\cite{deng2021voxel} (w/ \texttt{S.A.})  &  61.46 &  39.72 & 36.96 \\
            & Voxel-RCNN~\cite{deng2021voxel} (w/ \texttt{C.R.})  &  61.38 &  39.08 &  35.42  \\
            & Voxel-RCNN~\cite{deng2021voxel} (w/ \texttt{S.A.+C.R.})  & \textbf{62.31} & \textbf{40.74} & \textbf{40.63}  \\
            \toprule[1pt]
            \toprule[1pt]
            \multirow{2}{*}{only KITTI} & PV-RCNN~\cite{shi2020pv} (w/o \texttt{P.T.})  &  42.85  &  27.45  &  30.86 \\
            & PV-RCNN~\cite{shi2020pv} (w/ \texttt{P.T.} on nuScenes) &  44.38  &  28.09  &  31.44  \\
            \toprule[1pt]
            \multirow{2}{*}{only nuScenes} & PV-RCNN~\cite{shi2020pv} (w/o \texttt{P.T.})  &  38.82  &  23.82  &  4.40 \\
            & PV-RCNN~\cite{shi2020pv} (w/ \texttt{P.T} on KITTI) &  37.49  &  20.99  &  4.59   \\
            \toprule[1pt]
            \multirow{4}{*}{KITTI+nuScenes} & PV-RCNN~\cite{shi2020pv} (w/ \texttt{D.M.})  &  49.76  &  27.69  &  28.13 \\
            & PV-RCNN~\cite{shi2020pv} (w/ \texttt{C.A.})  & 51.24  &  24.59  & 21.49  \\
            & PV-RCNN~\cite{shi2020pv} (w/ \texttt{S.A.})  & 59.12  &  32.27  & 33.61 \\
            & PV-RCNN~\cite{shi2020pv} (w/ \texttt{C.R.})  & 62.44  &  37.37  & 34.13 \\
            & PV-RCNN~\cite{shi2020pv} (w/ \texttt{S.A.+C.R.})  & \textbf{63.58}  & \textbf{37.42} & \textbf{38.47}  \\
            \toprule[1pt]
        \end{tabular}
    }
    \end{small}
    \caption{Average (Avg.) detection results of joint training on KITTI and nuScenes datasets. The experiment and evaluation settings follow Table~\ref{tab:avg_waymo_nusc}.}
    \label{tab:avg_ki_nusc}
\end{table*}

\begin{table*}[t]
    \centering
    \begin{small}
    \resizebox{0.80\linewidth}{!}
    {
        \begin{tabular}{c|c|c|c|c}
            \bottomrule[1pt]
            \multirow{2}{*}{Trained on} & \multirow{2}{*}{Baseline Detectors}  & \multicolumn{3}{c}{Avg. on KITTI+Waymo}  \\
            \hhline{~~|-|-|-}
            &  & Car\&Vehicle & Pedestrian & Cyclist\\
            \hline
            \multirow{2}{*}{only KITTI} & Voxel-RCNN~\cite{deng2021voxel} (w/o \texttt{P.T.}) & 43.86 & 36.70 &  37.62  \\
            & Voxel-RCNN~\cite{deng2021voxel} (w/ \texttt{P.T.} on Waymo) & 45.06  &  38.12  & 36.61 \\
            \toprule[1pt]
            \multirow{2}{*}{only Waymo} & Voxel-RCNN~\cite{deng2021voxel} (w/o \texttt{P.T.})  &  47.44  &  68.55 &  59.69  \\
            & Voxel-RCNN~\cite{deng2021voxel} (w/ \texttt{P.T.} on KITTI) &  46.38  &  65.72  & 56.35  \\
            \toprule[1pt]
            \multirow{4}{*}{KITTI+Waymo} & Voxel-RCNN~\cite{deng2021voxel} (w/ \texttt{D.M.})  & 53.22  &  64.83  &  60.41  \\
            & Voxel-RCNN~\cite{deng2021voxel} (w/ \texttt{S.A.})  & \textbf{78.82}  &  65.65  &  65.92 \\
            & Voxel-RCNN~\cite{deng2021voxel} (w/ \texttt{C.R.})  & 78.14  &  63.06  &  65.27 \\
            & Voxel-RCNN~\cite{deng2021voxel} (w/ \texttt{S.A.+C.R.}) &  78.61 & \textbf{65.90} & \textbf{66.56} \\
            \toprule[1pt]
            \toprule[1pt]
            \multirow{2}{*}{only KITTI} & PV-RCNN~\cite{shi2020pv} (w/o \texttt{P.T.})  &  43.07 &  31.36 &  33.78  \\
            & PV-RCNN~\cite{shi2020pv} (w/ \texttt{P.T.} on Waymo) & 46.09  & 35.49  &  34.32  \\
            \toprule[1pt]
            \multirow{2}{*}{only Waymo} & PV-RCNN~\cite{shi2020pv} (w/o \texttt{P.T.}) & 64.89  &  65.24  &  57.32  \\
            & PV-RCNN~\cite{shi2020pv} (w/ \texttt{P.T.} on KITTI) & 48.50  & 63.02  & 56.44  \\
            \toprule[1pt]
            \multirow{4}{*}{KITTI+Waymo} & PV-RCNN~\cite{shi2020pv} (w/ \texttt{D.M.})  & 59.52 & 59.41 &  60.45 \\
            & PV-RCNN~\cite{shi2020pv} (w/ \texttt{S.A.})  & 78.69 & 61.77 & 64.71 \\
            & PV-RCNN~\cite{shi2020pv} (w/ \texttt{C.R.}) & 78.75 & 61.22 & 64.08 \\
            & PV-RCNN~\cite{shi2020pv} (w/ \texttt{S.A.+C.R.})  & \textbf{79.11} & \textbf{65.22} & \textbf{64.84} \\
            \toprule[1pt]
        \end{tabular}
    }
    \end{small}
    \caption{Average (Avg.) detection results of joint training on KITTI and Waymo datasets. The experiment and evaluation settings follow Table~\ref{tab:avg_waymo_nusc}.}
    \label{tab:avg_ki_wy}
\end{table*}

\begin{table*}[t]
    \centering
    \begin{small}
    \resizebox{0.99\linewidth}{!}
    {
        \begin{tabular}{c|c|c|c|c|c|c|c}
            \bottomrule[1pt]
            \multirow{2}{*}{Trained on} & \multirow{2}{*}{Baseline Detectors}  & \multicolumn{3}{c|}{Tested on Waymo} & \multicolumn{3}{c}{Tested on nuScenes} \\
            \hhline{~~|-|-|-|-|-|-}
            &  & Vehicle & Pedestrian & Cyclist & Car & Pedestrian & Cyclist \\
            \hline
            \multirow{2}{*}{Waymo+nuScenes} & Voxel-RCNN~\cite{deng2021voxel} (w/ \texttt{D.M.})  & 58.26 / 57.87  &  52.72 / 47.11  &  50.26 / 49.50   &   51.40 / 31.68  & 15.04 / 9.99   &   5.40 / 3.87  \\
            & Voxel-RCNN~\cite{deng2021voxel} (w/ \texttt{C.A.+S.A.+C.R.})  &  66.94 / 66.37 & 67.04 / 60.57  & 62.21 / 61.32  & 60.18 / 42.23  &  30.08 / 24.37  &  14.60 / 12.32  \\
            \toprule[1pt]
            \toprule[1pt]
            \multirow{2}{*}{Trained on} & \multirow{2}{*}{Baseline Detectors}  & \multicolumn{3}{c|}{Tested on Waymo} & \multicolumn{3}{c}{Tested on KITTI} \\
            \hhline{~~|-|-|-|-|-|-}
            &  & Vehicle & Pedestrian & Cyclist & Car & Pedestrian & Cyclist \\
            \hline
            \multirow{2}{*}{Waymo+KITTI} & Voxel-RCNN~\cite{deng2021voxel} (w/ \texttt{D.M.})  & 66.31 / 65.85  &  65.39 / 59.47  &  62.69 / 61.80 &  74.53 / 32.11  &  60.11 / 54.85  &   59.69 / 55.94  \\
            & Voxel-RCNN~\cite{deng2021voxel} (w/ \texttt{S.A.+C.R.})  &  66.73 / 66.26  & 66.39 / 60.36  &  64.52 / 63.55  &  90.03 / 82.39 &  62.51 / 57.01  &  69.52 / 66.30  \\
            \toprule[1pt]
        \end{tabular}
    }
    \end{small}
    \vspace{-0.3cm}
    \caption{Waymo-nuScenes and KITTI-Waymo dataset consolidation results, where the detection results on Waymo dataset are LEVEL\_2 evaluation metric.}
    \label{tab:level2}
\end{table*}

\setcounter{figure}{6}
\begin{figure*}
\centering
\includegraphics[width=14.0cm]{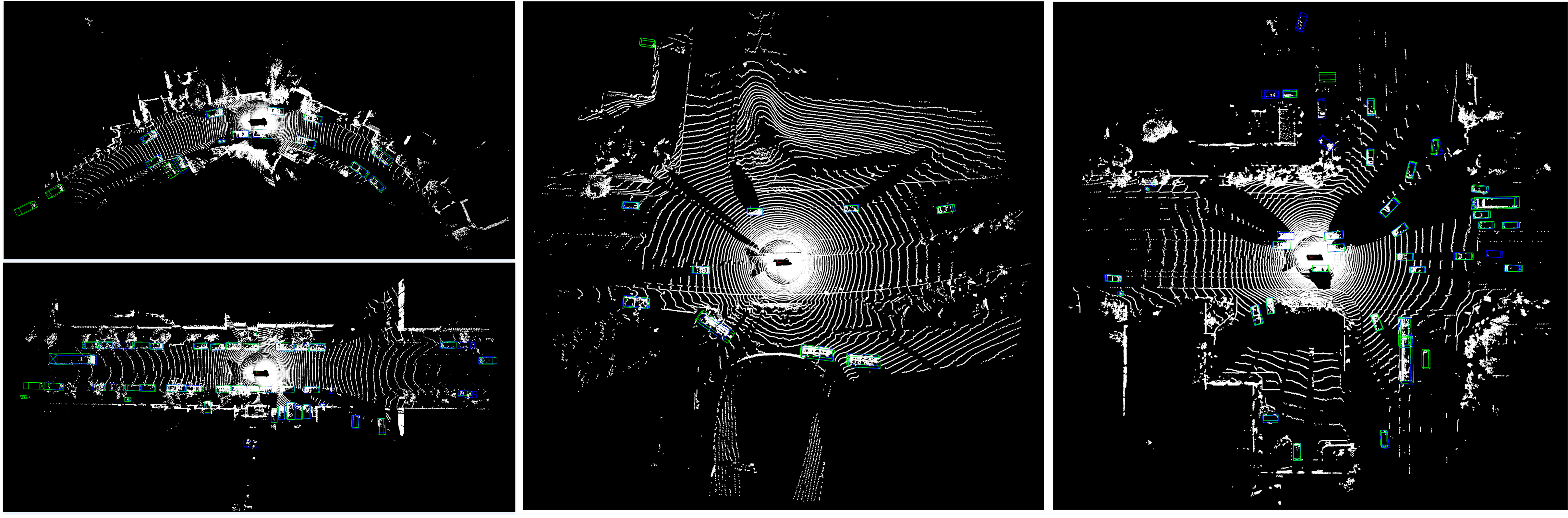}
\caption{Waymo detection results of jointly training on Waymo and nuScenes dataset, where the \textcolor{green}{green boxes} denote the 3D bounding boxes, while the \textcolor{blue}{blue boxes} represent the predicted 3D boxes.}
\label{fig:waymo}
\end{figure*}

\setcounter{figure}{7}
\begin{figure*}
\centering
\includegraphics[height=5.5cm, width=12.0cm]{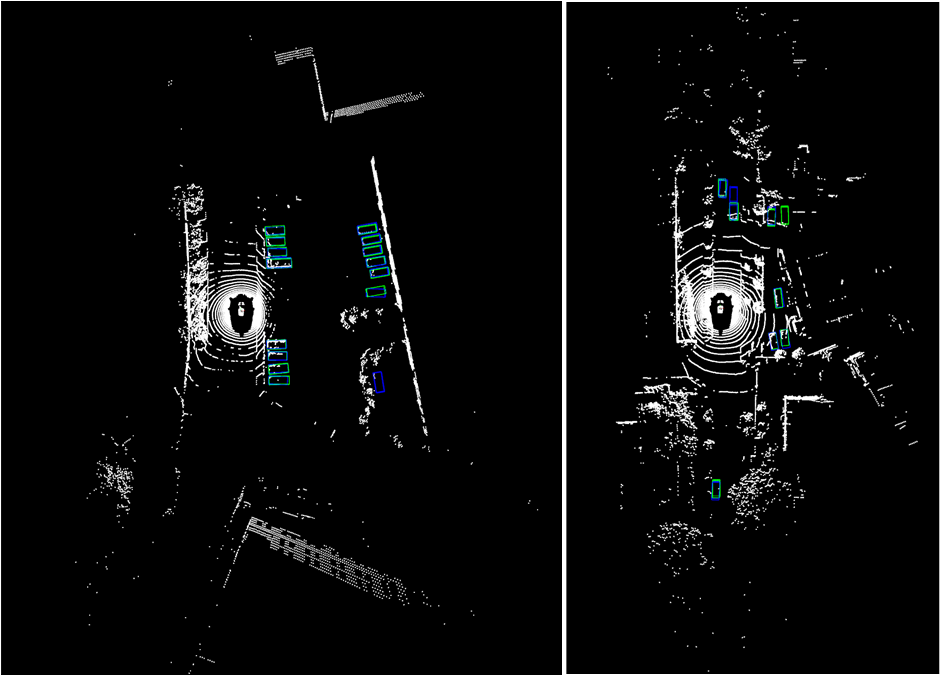}
\caption{nuScenes detection results of jointly training on Waymo and nuScenes dataset, where the \textcolor{green}{green boxes} denote the 3D bounding boxes, while the \textcolor{blue}{blue boxes} represent the predicted 3D boxes.}
\label{fig:nusc}
\end{figure*}

Originally, we have made a lot of attempts to train a baseline detector from multiple datasets, by directly merging the existing 3D datasets such as merging Waymo~\cite{sun2020scalability} and nuScenes~\cite{caesar2020nuscenes}. However, we found that, the commonly-used 3D baseline detection models such as PV-RCNN~\cite{shi2020pv} and Voxel-RCNN~\cite{mao2021voxel} suffer from severe detection performance degradation issue, when they are jointly trained on multiple 3D detection datasets. Similarly, previous 2D research works~\cite{zhou2022simple,wang2019towards} also point out that training a single 2D detector under multiple datasets together still faces a great challenge. But compared with 2D image-domain dataset-level consolidation, achieving such multi-dataset detection is more challenging in 3D point cloud scenario, which is mainly due to: \textbf{1) Data-level Differences}: Different datasets are often collected and constructed by different sensor types, and \textbf{2) Taxonomy Differences}: Different datasets constructed by multi-manufacturers often present inconsistent class-label definition.

Actually, we found that for 3D autonomous driving scenes, a major reason of the dataset-interference is that the above-mentioned data-level differences between different datasets are huge, as illustrated in Fig~\ref{fig:compare}. For example, the point cloud range is very inconsistent between different datasets, which results in the voxel-wise receptive field size of different datasets being different under the single-dataset training paradigm. Further, as shown in Table~\ref{tab:range}, employing the original point range of each dataset to train a detector cannot achieve a good generalization ability, compared with that we align the point range from different datasets.

\subsection{Uni3D Inference Usage}
\label{sec:experiments}
In the main text, many options for achieving the single-dataset inference by Uni3D are considered and studied, and we report the class-wise detection results for each individual dataset. Here, we further deeply analyze the differences between different single-dataset inference methods as illustrated in Tables~\ref{tab:inference} and~\ref{tab:ensembling}.

\noindent \textbf{Ablation Studies on Single-dataset Inference using Uni3D.} In this part, we try to deeply investigate the \texttt{C.R.} module in Uni3D from the following two aspects:

\noindent \noindent 1) \underline{Inference usage for \texttt{C.R.} module}:
As described in the main text, the purpose of the \texttt{C.R.} module is to exploit the reusable features \textit{during the model training stage}. However, exploring such feature relations across datasets will cause the inconsistency between the model multi-dataset training and single-dataset testing, mainly due to that the shared features {\small $\hat{f}^{bev}_{shared}$} are dependent on multiple inputs. In this part, we provide many options of Uni3D during the inference usage, and show experimental comparison results. For option one, \textbf{BEV feature copy} method indicates that during the inference stage, BEV features from the single dataset will be simultaneously copied to $f_i^{bev}$ and $f_j^{bev}$, in order to obtain the shared features {\small $\hat{f}^{bev}_{shared}$}. And then, the dataset-specific BEV features can be obtained using the dataset-specific SE module, which uses the {\small $\hat{f}^{bev}_{shared}$} as the input. For option two, \textbf{BEV feature mask} means that we set the input BEV tensor (features) of the other dataset to zero, and produce the shared features {\small $\hat{f}^{bev}_{shared}$}. 

The corresponding experimental results are shown in Table~\ref{tab:inference}. We observe that, when performing the single-dataset model inference, the \textbf{BEV feature copy} method achieves better detection accuracy. This is mainly because Uni3D introduces the BEV feature interaction across different datasets during the multi-dataset training process. However, the \textbf{BEV feature mask} method will mask the BEV features from one of the two branch, significantly increasing the distribution differences of BEV features when performing the inference (a zero tensor \textit{v.s.} BEV features from point cloud data). Actually, although the BEV feature copy method can ensure the Uni3D training-and-testing consistency, such a method is not the optimal solution to tackle the inconsistency issue between multi-dataset training and single-dataset inference when fusing the BEV features from different datasets during the model training stage, which will be a possible research topic in our future work.

\noindent 2) \underline{The comparisons against model ensemble}: Further, in order to verify the effectiveness of fusing BEV features from different datasets during the model training stage, we compare our Uni3D with the model ensemble method. Specifically, in order to avoid the impact of data-level correction operation, we first use the baseline model with point range alignment and statistics-level alignment (\textit{i.e.} Voxel-RCNN (w/ \texttt{C.A.+S.A.}) baseline), and then fine-tune the detection head on only a Waymo dataset or a nuScenes dataset without using \texttt{C.R.} module, where the purpose of fine-tuning is to enable the model to perform the detection task on different datasets using the trained dataset-specific detection head. Finally, we perform the test time ensemble using the two dataset-specific detection heads trained on different datasets. The experimental results of test time ensemble are illustrated in Table~\ref{tab:ensembling}. We observe that the detection accuracy achieved by such dataset-level ensemble method is not satisfactory. This is due to that, one of the two ensemble heads (detection head) is trained only using nuScenes, and has strong prediction bias when testing on Waymo (\textit{e.g.}, \underline{only $67.32\%$ and $70.02\%$ AP} on Vehicle and Pedestrian).

\subsection{Ablation Studies of \texttt{C.R.} module}
In this part, we conduct the experiments of removing the attention component and SE component in \texttt{C.R.} module using two different baseline detectors. From Table~\ref{tab:at_se}, it can be seen that each newly-added component (the attention and SE) can bring accuracy gains on Waymo dataset.

\subsection{More Experimental Results}

In this part, we report the average detection accuracy across datasets as shown in Tables~\ref{tab:avg_waymo_nusc},~\ref{tab:avg_ki_nusc}, and~\ref{tab:avg_ki_wy}. And then, we give the detection results using LEVEL\_2 metric of Waymo-related experiments in Table~\ref{tab:level2}.

\noindent \textbf{Average Detection Results on Different Datasets.}
Following the experimental setting in the main text, we show the average detection results in order to better demonstrate the advantages of the proposed Uni3D. It should be emphasized that for different datasets, we employ different evaluation metrics such as 3D-AP/3D-APH on Waymo and BEV-AP/3D-AP nuScenes. Thus, in order to calculate the cross-dataset average detection accuracy, we simply average the 3D-AP results on Waymo dataset and 3D-AP results on KITTI or nuScenes dataset, which is conducive to keeping consistent with the results reported in the main text.

The corresponding the cross-dataset average results are shown in Table~\ref{tab:avg_waymo_nusc} to ~\ref{tab:avg_ki_wy}. It can be seen from Tables~\ref{tab:avg_waymo_nusc},~\ref{tab:avg_ki_nusc}, and~\ref{tab:avg_ki_wy} that, Uni3D is beneficial to boost the multi-dataset generalization ability of the existing 3D detection baseline (\textit{e.g.} PV-RCNN~\cite{shi2020pv} and Voxel-RCNN~\cite{mao2021voxel}), outperforming all the designed baselines in terms of average detection accuracy with a large margin. As a result, it can be concluded that it is feasible to just utilize a simple-and-versatile method to improve the dataset-level generalization ability of the existing 3D detection models, which further shows the great potential of the proposed method in future autonomous driving-related applications.

\noindent \textbf{LEVEL\_2 Detection Results on Different Waymo-related Experiments.} Here, Table~\ref{tab:level2} shows the LEVEL\_2 detection accuracy achieved by the proposed Uni3D. From Table~\ref{tab:level2}, we observe that compared with the \texttt{D.M.(Direct Merging)} baseline, our method achieves relatively high results for both Waymo and nuScenes or KITTI dataset.

\subsection{Detection Visualization Results}
\label{sec:qualitative_results}

More visualization results of the proposed Uni3D are shown in Figs.~\ref{fig:waymo} and~\ref{fig:nusc}.

For Figs.~\ref{fig:waymo} and~\ref{fig:nusc}, we use the proposed Uni3D jointly trained on Waymo and nuScenes datasets, and show the results on the validation set of Waymo and nuScenes. These results comprehensively demonstrate that we can achieve better detection results simultaneously for Waymo and nuScenes datasets only using a single detection model.


\clearpage
\clearpage


\end{document}